\documentclass[10pt,twocolumn,letterpaper]{article}

\usepackage[pagenumbers]{cvpr} 
\usepackage{graphicx}
\usepackage{multirow}
\usepackage{makecell}
\usepackage{booktabs}
\usepackage{colortbl} 
\usepackage{amssymb}
\usepackage{bbding}

\definecolor{ForestGreen}{RGB}{34,139,34}
\definecolor{dino}{RGB}{249,231,227}
\definecolor{aliceblue}{rgb}{0.94, 0.97, 1.0}

\definecolor{cvprblue}{rgb}{0.21,0.49,0.74}
\usepackage[pagebackref,breaklinks,colorlinks,allcolors=cvprblue]{hyperref}

\def\confName{CVPR}
\def\confYear{2025}

\title{CPath-Omni: A Unified Multimodal Foundation Model for Patch and Whole Slide Image Analysis in Computational Pathology}

\def\@fnsymbol#1{\ensuremath{%
		\ifcase#1
		\or 
		\dagger
		\or 
		\ddagger
		\or 
		\mathsection
		\or 
		\mathparagraph
		\else 
		\@ctrerr  
		\fi}}   
\makeatother 
\author{
	$^{1,2}$Yuxuan Sun\footnotemark[1]\;,
        $^{1}$Yixuan Si\thanks{Equal contribution.}\;,
	$^{1}$Chenglu Zhu,
	$^{3}$Xuan Gong,
	$^{4}$Kai Zhang,
	$^{1,2}$Pingyi Chen, \\
	$^{5}$Ye Zhang,
	$^{1,2}$Zhongyi Shui,
	$^{1}$Tao Lin\footnotemark[2]\;, 
	$^{1}$Lin Yang\thanks{Corresponding author.} \vspace{1mm}\\
    $^1$Westlake University, $^2$Zhejiang University, $^3$Harvard University, \\
    $^4$The Ohio State University, 
    $^5$University of the Chinese Academy of Sciences}

\begin{document}
\maketitle
\begin{abstract}
The emergence of large multimodal models (LMMs) has brought significant advancements to pathology. Previous research has primarily focused on separately training patch-level and whole-slide image (WSI)-level models, limiting the integration of learned knowledge across patches and WSIs, and resulting in redundant models. In this work, we introduce CPath-Omni, the first 15-billion-parameter LMM designed to unify both patch and WSI level image analysis, consolidating a variety of tasks at both levels, including classification, visual question answering, captioning, and visual referring prompting.
Extensive experiments demonstrate that CPath-Omni achieves state-of-the-art (SOTA) performance across seven diverse tasks on 39 out of 42 datasets, outperforming or matching task-specific models trained for individual tasks. Additionally, we develop a specialized pathology CLIP-based visual processor for CPath-Omni, CPath-CLIP, which, for the first time, integrates different vision models and incorporates a large language model as a text encoder to build a more powerful CLIP model, which achieves SOTA performance on nine zero-shot and four few-shot datasets. Our findings highlight CPath-Omni's ability to unify diverse pathology tasks, demonstrating its potential to streamline and advance the field of foundation model in pathology.
\end{abstract}    
\section{Introduction}
\label{sec:intro}

\begin{figure}[t]
	\centering
	\includegraphics[width=\linewidth]{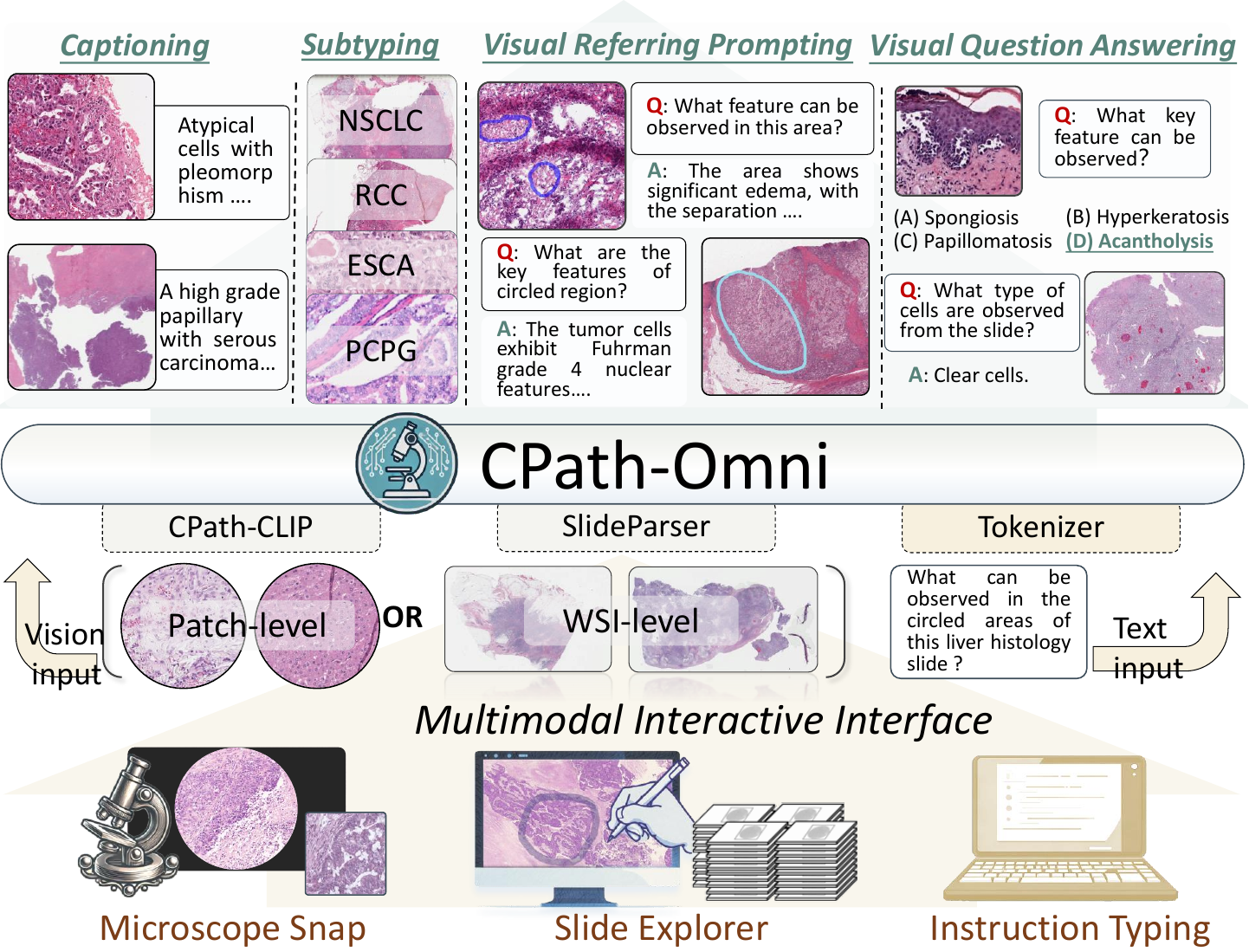}
	
	\caption{Overview of CPath-Omni’s ability to handle both patch-level and WSI analysis in clinical environments, such as microscope views and scanned WSIs, while supporting various tasks.}
	\label{fig:overall_structral}
\end{figure}

Pathology plays a pivotal role in modern medicine, serving as the foundation for diagnosing and understanding diseases \cite{kumar2014robbins}. However, pathology requires significant human effort to conduct precise and accurate interpretations of images that can be as large as 100,000 $\times$ 100,000 pixels. 

In recent years, with advancements in computational power and the digitization of pathology, a wide range of models have been developed to assist pathologists in their diagnostic tasks. These include CLIP \cite{radford2021learning}-based patch processing models like CONCH \cite{lu2024visual} and PathGen-CLIP \cite{sun2024pathgen}, DINOv2 \cite{oquab2023dinov2}-based models like Virchow2 \cite{zimmermann2024virchow} and UNI \cite{chen2024towards}, and LMMs like PathAsst \cite{sun2024pathasst}, PathGen-LLaVA \cite{sun2024pathgen}, Quilt-LLaVA \cite{seyfioglu2024quilt}, and PathChat \cite{lu2024multimodal}, which support tasks such as multi-turn conversations. At the WSI level, models like Prov-GigaPath \cite{xu2024whole} and HIPT \cite{chen2022scaling} are developed for WSI classification, while models such as HistGen \cite{guo2024histgen} and WsiCaption \cite{chen2024wsicaption} are used to generate WSI reports.

In this work, we propose CPath-Omni, a multimodal foundation model designed to unify patch-level and WSI-level analysis. CPath-Omni can perform diverse tasks such as VQA, classification, captioning, and visual referring prompting. By integrating these two levels of analysis and enabling generalizable task performance, CPath-Omni represents a significant step toward developing a truly versatile and comprehensive assistive tool for pathologists.

We first train a novel pathology-specific foundation model, CPath-CLIP, to serve as the vision encoder for CPath-Omni. CPath-CLIP is the first model to integrate a large language model (LLM) as the text encoder for CLIP, while also incorporating the self-supervised pathology vision model Virchow2 as the visual encoder, alongside the original CLIP-L model. To train CPath-CLIP, we collect 700,145 image-caption pairs from diverse datasets, constructing CPath-PatchCaption. Then, we integrate CPath-CLIP into the large language model Qwen2.5-14B \cite{hui2024qwen2} to equip it with visual capabilities, creating the CPath-Omni.

The training of CPath-Omni proceeds through four stages to build a unified model capable of handling both patch-level and WSI-level tasks. In the first stage, we pre-align CPath-CLIP with the Qwen2.5-14B language model using the CPath-PatchCaption dataset. Next, we collect and construct 351,871 instruction tuning samples from four diverse patch-level tasks across 21 datasets, including patch-level classification, VQA, captioning, and visual referring prompting.
In stage 3, we introduce WSI-related data, including 5,850 cleaned WSI reports, to continue pretraining CPath-Omni, further enhancing its WSI understanding based on the previous stage. In the final stage, we construct 33,830 WSI instruction tuning samples from three tasks across nine datasets, including classification, VQA, and captioning, along with 15\% patch-level instruction tuning samples for joint WSI-Patch training.
This joint training enables CPath-Omni to seamlessly process both patch and WSI data and enables a wide range of downstream tasks.

Extensive experiments across seven diverse tasks and 42 datasets are conducted to validate the effectiveness of CPath-Omni. With its broad capabilities, CPath-Omni achieves state-of-the-art (SOTA) performance on 39 out of 42 datasets and demonstrates comparable or superior performance to task-specific models. The main contributions of our study are summarized as follows:

\begin{itemize} 
\item We develop \textbf{CPath-CLIP}, the most powerful pathology CLIP model to date, which achieves SOTA results on 9 zero-shot and 4 linear probing classification datasets. 
\item We introduce \textbf{CPath-Omni}, the first unified model capable of handling both patch-level and WSI analysis across diverse tasks, offering exceptional performance and versatility, and representing an early realization of the ``one-for-all" paradigm in computational pathology. 
\item We curate a diverse and comprehensive training and testing dataset, spanning 7 tasks across 42 datasets, making it the largest and most diverse dataset for training LMMs in pathology. Extensive experiments are conducted on these datasets to confirm CPath-Omni's significant advancement in pathology foundation models.
\end{itemize}                                                                                                                                                                                                                                                                                                             
\section{Related Work}

\textbf{Vision Foundation Models in Pathology.} 
In recent years, with the rapid advancement of digital pathology slide digitization, pathology-specific visual foundation models (VFMs) have made significant strides. These models are primarily divided into two main categories. The first is the Vision-Language-based model, such as CLIP \cite{radford2021learning}, which employs contrastive learning to align images with textual descriptions, enabling the vision encoder to generate semantically meaningful features. Researchers have compiled large datasets of image-caption pairs from sources such as PubMed, YouTube, Twitter, and books to train these models. Notable examples in this category include Quilt-Net \cite{ikezogwo2024quilt}, PLIP \cite{huang2023visual}, PathCLIP \cite{sun2024pathasst}, PathGen-CLIP \cite{sun2024pathgen}, and CONCH \cite{lu2024visual}. The second category focuses on vision-only models, trained through self-supervised learning using vast amounts of patch data extracted from WSIs. These models are typically trained with techniques like DINO \cite{caron2021emerging, oquab2023dinov2} pretraining to learn robust visual representations. Prominent models in this group include Lunit \cite{kang2023benchmarking}, UNI \cite{chen2024towards}, Prov-GigaPath \cite{xu2024whole}, and Virchow series \cite{vorontsov2023virchow, zimmermann2024virchow}.

The development of pathology-specific VFMs has significantly improved image representations, enhancing performance on downstream tasks like patch and WSI classification. CLIP-based models, which are pre-aligned with textual information, are easier to integrate with LLMs. In contrast, DINO-based models tend to learn more fine-grained visual features \cite{jiang2023clip}. 
In this paper, we combine the strengths of both approaches by leveraging OpenAI-CLIP-L and the vision-only Virchow2 as our visual encoder, aligned with the Qwen2-1.5B \cite{yang2024qwen2} LLM to enhance visual capabilities and improve alignment with LLM world knowledge.

\textbf{Multimodal Generative Foundation Models in Pathology.} 
The integration of LLMs like GPT-4~\cite{gpt4} with vision capabilities has led to advanced LMMs such as GPT-4V~\cite{openai2023gpt4v} and Gemini Pro Vision~\cite{team2023gemini}. These LMMs offer generalized capabilities, which are particularly well-suited to the field of pathology, where understanding a broad range of diseases (e.g., lung cancer, liver cancer), adapting to various tissues (e.g., prostate, colon, stomach), and performing across diverse tasks (e.g., tumor classification, survival prediction) is essential. As a result, numerous pathology-specific LMMs have been developed, building on these general LLMs and LMMs. Notable models include PathAsst \cite{sun2024pathasst}, PathGen-LLaVA \cite{sun2024pathgen}, Quilt-LLaVA \cite{seyfioglu2023quilt}, and PathChat \cite{lu2024multimodal}, which demonstrate strong image understanding and multi-turn conversational abilities. However, due to input size constraints, these models primarily focus on patch-level tasks.

Recently, several works have trained smaller multimodal language models for WSI tasks. For example, WsiCaption \cite{chen2024wsicaption} and HistGen  \cite{guo2024histgen} focus on WSI caption generation, while WSI-VQA \cite{chen2025wsi} targets WSI-based VQA. 
The former primarily uses publicly available datasets with fewer than 10,000 samples. More recently, The PRISM \cite{shaikovski2024prism}, trained on 587,196 internal WSIs, developed a CoCa \cite{yu2022coca}-like model capable of zero-shot WSI classification and WSI report generation, representing a significant step toward to more generalizable generative foundation models.

\textbf{Multimodal Datasets in Pathology.}
To construct powerful CLIP-based models, a large volume of high-quality image-caption pairs is essential. In the patch-level domain, the ARCH \cite{gamper2021multiple} dataset collects 8,617 figure-caption pairs related to histology images from medical articles and textbooks. The PathCap~\cite{sun2024pathasst} dataset contains 207,000 pathology image-caption pairs, carefully curated from over 15 million image-text pairs sourced from PubMed and various textbooks. The OpenPath~\cite{huang2023visual} dataset includes 208,414 pairs collected from Twitter posts, while the QUILT-1M~\cite{ikezogwo2024quilt} dataset contains 768,826 histopathology image-text pairs derived from YouTube video frames. Additionally, to train LMMs, Quilt-Instruct \cite{seyfioglu2023quilt} generated 107,131 instruction-tuning samples from YouTube lectures, while PathGen-Instruct \cite{sun2024pathgen} created 200K instruction-tuning samples based on synthetic captions from the PathGen-1.6M.

In the WSI domain, the available data is much more limited. WsiCaption \cite{chen2024wsicaption} and HistGen  \cite{guo2024histgen} generate 10,000 and 7,753 WSI captioning samples based on TCGA report PDFs, respectively. More recently, WSI-VQA \cite{chen2025wsi} expanded these WSI report datasets to create the WSI VQA dataset.

In this paper, we systematically compile these existing datasets and augment them with additional processing. We also incorporate our downstream datasets as the foundation for the training and testing of CPath-Omni.
\section{Data Preparation}

In this section, we introduce the patch-level and WSI-level data required for constructing CPath-Omni.
\subsection{Patch Level Dataset}
\textbf{CPath-PatchCaption}: This dataset is a curated image-caption pairs dataset consisting of 700,145 pairs gathered from various open-source datasets. Specifically, it includes 218,630 pairs from PathCap, 388,932 pairs from Quilt-1M, and 92,583 pairs from OpenPath. We ensured that this caption dataset does not overlap with the test data used in Path-Omni. This dataset serves as a key component for pretraining CPath-CLIP and the stage 1 pretraining of CPath-Omni.

\noindent\textbf{CPath-PatchInstruction}: CPath-PatchInstruction is a diverse dataset consisting of 351,871 samples across captioning, VQA, classification, and visual referring prompting tasks. Of these, 147,843 examples are sourced from CPath-PatchCaption, representing the highest-quality samples curated by human annotators. Captions and images were further refined using GPT-4 to generate more detailed captions, enhancing CPath-Omni's powerful image understanding capabilities. Additionally, 40,000 examples were drawn from PathInstruct, a multimodal, multi-turn conversational dataset.
For the classification data, samples were collected from various public classification datasets, including VALSET-TCGA, VALSET-WNS, VALSET-CHA \cite{tolkach2023artificial}, Stomach, KIRC, CocaHis~\cite{SITNIK2021102402}, PAIP23, BNCB~\cite{xu2021predicting}, CATCH~\cite{Wilm2022CanineCATCH}, PAIP21, MIDOG22~\cite{MIDDog-dataset}, KICH, CAMEL~\cite{9008367}, Gleason-CNN~\cite{DVN/OCYCMP_2018}, OCELOT~\cite{Ryu_2023_CVPR}, and NASNetLarge~\cite{zoph2018learning}. No more than 5,000 data points were randomly sampled from each dataset, with 80\% converted into VQA format for CPath-Instruction training and 20\% reserved for validation and testing.   
To further enhance the model's capabilities and interpretability, we collaborate with expert pathologists who annotate 1,300 high-resolution images randomly selected from TCGA. The annotations are initially processed by GPT-4 to generate preliminary findings of the images, which are then meticulously revised, supplemented, and refined by pathologists for accuracy. Most importantly, corresponding areas for each pathology finding are highlighted in the images to create visual referring prompting data. Of these, 1,200 data points are used for training, and 100 are designated for validation and testing. The entire CPath-PathInstruction dataset is used for stage 2 training of CPath-Omni. For further construction and annotation details, please refer to the Appendix.

\subsection{WSI Level Dataset}

\textbf{CPath-WSIInstruction}: CPath-WSIInstruction encompasses captioning, VQA, and classification data at the WSI level. The dataset includes 7,312 WSI-level captioning examples sourced from HistGen. To ensure that the validation and test sets do not overlap with the WSI classification data, we re-divide the data into training, validation, and test sets with an 8:1:1 ratio. The training set is incorporated into CPath-WSIInstruction. For the VQA component, we further generate a WSI VQA dataset by prompting GPT-4 based on the divided WSI captions. Specific prompts are detailed in the Appendix.
For classification, we compile subtype data for 8 TCGA subtyping tasks, including RCC, NSCLC, BRCA, UCEC, THCA, ESCA, BLCA, and TGCT. 80\% of this data is transformed into the QA format that CPath-Omni can accept for training, while the remaining 10\% is reserved for validation and testing.
\section{The Proposed CPath-Omni}
As shown in \cref{fig:overall_structral}, CPath-Omni's architecture consists of two vision components, CPath-CLIP (for patch-level image processing) and SlideParser (for WSI-level image processing), along with an LLM. Patch and WSI inputs are processed through their respective branches. These vision modules encode patch-level or WSI-level visual tokens, which are then fed as input to the LLM. For the LLM, we choose the latest Qwen2.5 14B model as part of CPath-Omni. In this section, we provide a detailed description of the construction of CPath-CLIP and SlideParser.

\begin{figure*}[t]
	\centering
	\includegraphics[width=\linewidth]{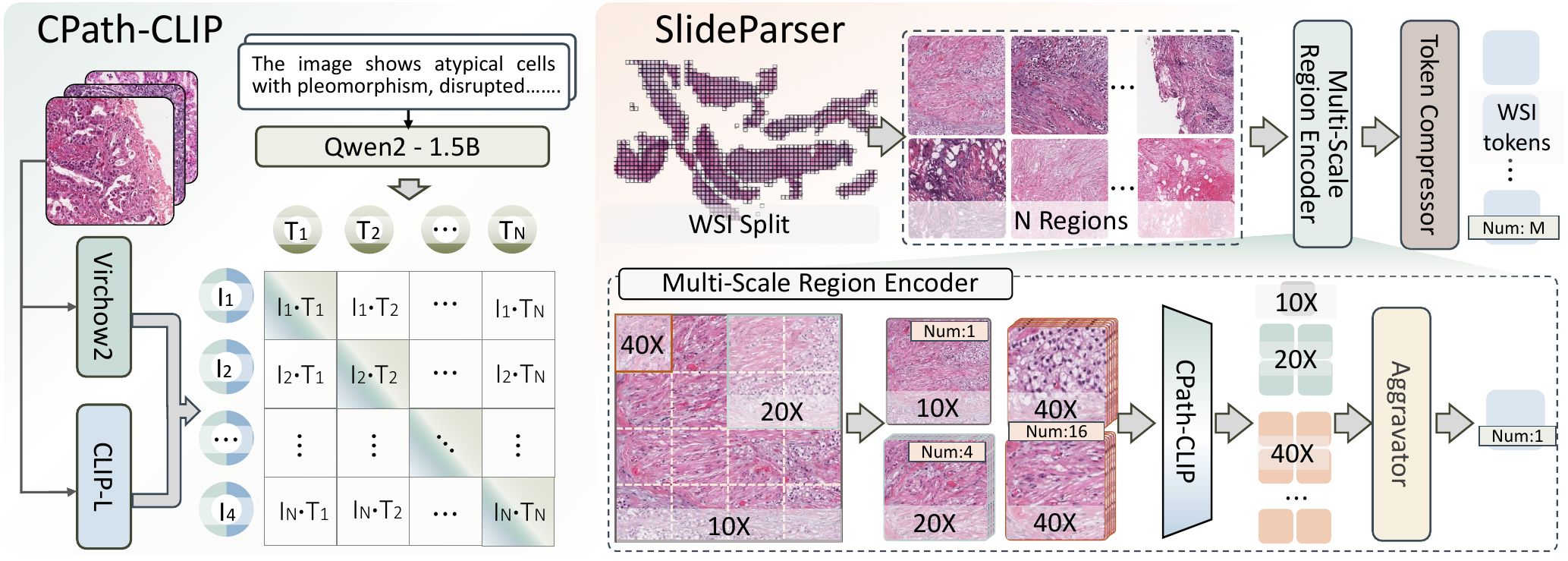}
	
	\caption{Overview of two key vision components of CPath-Omni: the patch-level model, CPath-CLIP, and the WSI model, SlideParser.}
	\label{fig:CPath-CLIP_SlideParser}
\end{figure*}

\subsection{CPath-CLIP}

Pretrained CLIP models are commonly used with LLMs in general-domain LMMs, as they are well-aligned with the text space and are easier to integrate with LMMs. However, pathology images differ significantly from natural images, creating a domain gap that limits CLIP performance in pathology tasks. Therefore, a pathology-specific CLIP-based model should be constructed to improve image understanding and enhance LMM performance in this domain.

One challenge with CLIP models is their focus on coarse-grained semantic information, which can cause the loss of fine-grained details critical for pathology diagnoses, such as chromatin structure or mitosis. To address this, we integrate the pretrained Virchow2 model, trained on 3 million WSIs using DINOv2-based vision pretraining, providing strong visual representations. We also retain OpenAI's CLIP vision component to preserve robust semantic features. As shown in the left part of \cref{fig:CPath-CLIP_SlideParser}, we feed the image into both models and concatenate their feature outputs. 
To further improve the alignment between the vision model and the LLM, we replace the GPT-2 model used in the original CLIP architecture with Qwen2-1.5B, a model from the same framework as the LLM. Due to its larger size and broader training corpus, Qwen2-1.5B brings stronger world knowledge, significantly improving the semantic alignment between the vision and language components. Specifically, we train CPath-CLIP using the CPath-PathCaption dataset based on OpenCLIP framework \cite{ilharco_gabriel_2021_5143773}.

When used for patch processing in CPath-Omni, CPath-CLIP adopts the AnyRes strategy from Llava-Next \cite{liu2024llavanext} to handle higher resolutions. Each image is split into a 3$\times$3 grid of sub-patches, which are encoded by CPath-CLIP to extract features. These features are then processed through a two-layer MLP before being input into the LLM.

\subsection{SlideParser}

SlideParser is the core component for handling WSI inputs, particularly for gigapixel images up to 100,000 $\times$ 100,000 pixels. To manage this, we first split the WSI into multiple 2048 $\times$ 2048 image regions. As pathologists often require both global and local context during analysis, we incorporate a multi-scale region encoding approach. As shown in \cref{fig:CPath-CLIP_SlideParser}. Each 2048 $\times$ 2048 region is further subdivided into three scales: 16 tiles of 512 $\times$ 512, 4 tiles of 1024 $\times$ 1024, and 1 tile of 2048 $\times$ 2048. These tiles are then encoded by CPath-CLIP to generate image-level features, which are aggregated using average pooling to produce a final multi-scale feature representation for the 2048 $\times$ 2048 region, which also significantly reduces the number of image tokens input into the LLM.

Since WSIs can vary greatly in size—from dozens to thousands of patches—this variability can cause instability during LMM training. To address this, we introduce a token compression layer that standardizes the input by reducing WSI tokens to a fixed length. Specifically, we adopt the CoCa approach \cite{yu2022coca}, using 1152 query tokens through multi-head attention to query the patch tokens, resulting in a unified output of 1152 tokens for the LLM.

\subsection{The Training  of CPath-Omni}

The construction of CPath-Omni undergoes four training stages: patch-based pretraining, finetuning, WSI-based pretraining, and mixed patch-WSI training.

\textbf{Stage 1:} In this stage, we focus on aligning the feature spaces of CPath-CLIP and the LLM. Only the two-layer MLP connecting CPath-CLIP to the LLM is trained, using the CPath-PatchCaption dataset. This enables the model to pre-align visual and language features.

\textbf{Stage 2:} We unfreeze all model parameters and fine-tune using the CPath-PatchInstruct dataset. Building on the alignment from Stage 1, this stage enables CPath-Omni to learn a variety of tasks, including VQA, image classification, captioning, and pathology-related knowledge.

\textbf{Stage 3:} In this stage, training uses WSI report-only data. Only SlideParser is unfrozen to align WSI features with the pathology-related knowledge already learned by the LLM from patch-based training.

\textbf{Stage 4:} The final stage introduces mixed training with 15\% randomly sampled CPath-PathInstruct and CPath-WSIInstruct datasets. By this point, CPath-Omni has gained a strong understanding of pathology from previous stages, enabling effective transfer of patch-based knowledge to WSI tasks, despite the limited WSI data.

After completing these four stages, CPath-Omni is fully equipped to effectively handle both patch-based and WSI analysis across a variety of downstream tasks.

\begin{table*}[t!]
	\resizebox{\linewidth}{!}{
		\begin{tabular}{@{}ccccccccccc@{}}
			\toprule
			\textbf{Model}                    & \textbf{LC-Lung} & \textbf{LC-Colon} & \textbf{CRC100K} & \textbf{SkinCancer} & \textbf{Pcam} & \textbf{BACH} & \textbf{Osteo} & \textbf{WSSSLUAD} & \textbf{SICAPv2} & \textbf{Average} \\ \midrule
			OpenAI-CLIP-L              & 70.4    & 81.1     & 40.3    & 19.4        & 55.5 & 34.3 & 53.9  & 81.2     & 25.4    & 51.3 \\
			PLIP                     & 87.9      & 90.2       & 52.8      & 42.5         & 51.8   & 34.3   & 52.9    & 73.1     & 42.5      & 58.6 \\
			QuiltNet                & 80.0      & 91.0       & 49.5      & 46.4         & 58.7   & 43.8   & 53.8    & 70.5     & 37.3      & 58.9 \\
			PathCLIP        & 88.9    & 94.3   & 55.3    & 35.1        & 72.5 & 46.8   & 69.2   & \underline{85.1}     & 48.3    & 66.2 \\
			BiomedCLIP               & 48.8      & 94.3       & 29.9      & 31.7         & 84.0   & 39.8   & 36.7    & 73.7      & 32.2     & 52.9 \\
			
			PathGen-CLIP-L      & \underline{89.8}    & \underline{99.3}   & \textbf{78.0}    & \underline{70.6}        & \underline{88.2} & \underline{71.5}   & \underline{74.6}   & 82.2     & \textbf{63.5}    & \underline{79.7} \\
			\rowcolor{aliceblue} CPath-CLIP      & \textbf{97.1}    & \textbf{100.0}   & \textbf{78.0}    & \textbf{74.2}        & \textbf{95.9} & \textbf{72.3}   & \textbf{80.7}   & \textbf{87.1}     & \underline{63.1}    & \textbf{83.2} \\
			\bottomrule
	\end{tabular}}
	\caption{Zero-shot classification comparison of various CLIP models on different pathology image classification datasets with accuracy (\%). The best performance is highlighted in \textbf{bold}, while the second-best is \underline{underlined}.}
	\label{tab:zero-shot}
\end{table*}

\section{Experiments}
We conducted extensive experiments to evaluate the universal task-solving capabilities of CPath-Omni. At the patch level, we evaluated across 32 subsets, covering tasks such as VQA, classification, captioning, and visual referring prompting (VPR). For the WSI level, we evaluated 10 subsets focusing on WSI VQA, classification, and captioning.

\subsection{Benchmarking CPath-CLIP}
In our experiments, we evaluated the image-text alignment and feature extraction capabilities of CPath-CLIP through zero-shot classification and few-shot linear probing. For zero-shot classification, we utilized datasets including PatchCamelyon (Pcam)~\cite{veeling2018rotation}, CRC-100K~\cite{kather2018100}, SICAPv2~\cite{silva2020going}, BACH~\cite{aresta2019bach}, Osteo~\cite{arunachalam2019viable}, SkinCancer~\cite{kriegsmann2022deep}, WSSSLUAD~\cite{han2022wsss4luad}, LC-Lung, and LC-Colon~\cite{borkowski2019lung}. Our model's performance was benchmarked against OpenAI-CLIP-L~\cite{radford2021learningtransferablevisualmodels}, PLIP~\cite{huang2023visual}, QuiltNet~\cite{ikezogwo2023quilt1m}, PathCLIP~\cite{sun2024pathasst}, BiomedCLIP~\cite{zhang2023biomedclip}, and the SOTA PathGen-CLIP-L~\cite{sun2024pathgen}.

\noindent\textit{\textbf{Results: CPath-CLIP achieved superior performance across most datasets.}} As shown in \cref{tab:zero-shot}, CPath-CLIP notably surpasses the current SOTA model, PathGen-CLIP-L by a significant margin on the Osteo, Pcam, and LC-Lung datasets, with improvements of 6.1\%, 7.7\% and 7.3\%, respectively. Additionally, it demonstrated significant advantages over other models, underscoring CPath-CLIP’s enhanced image-text feature alignment capabilities driven by its stronger vision and language model integration.

\begin{figure*}[t]
	\centering
	\includegraphics[width=\linewidth]{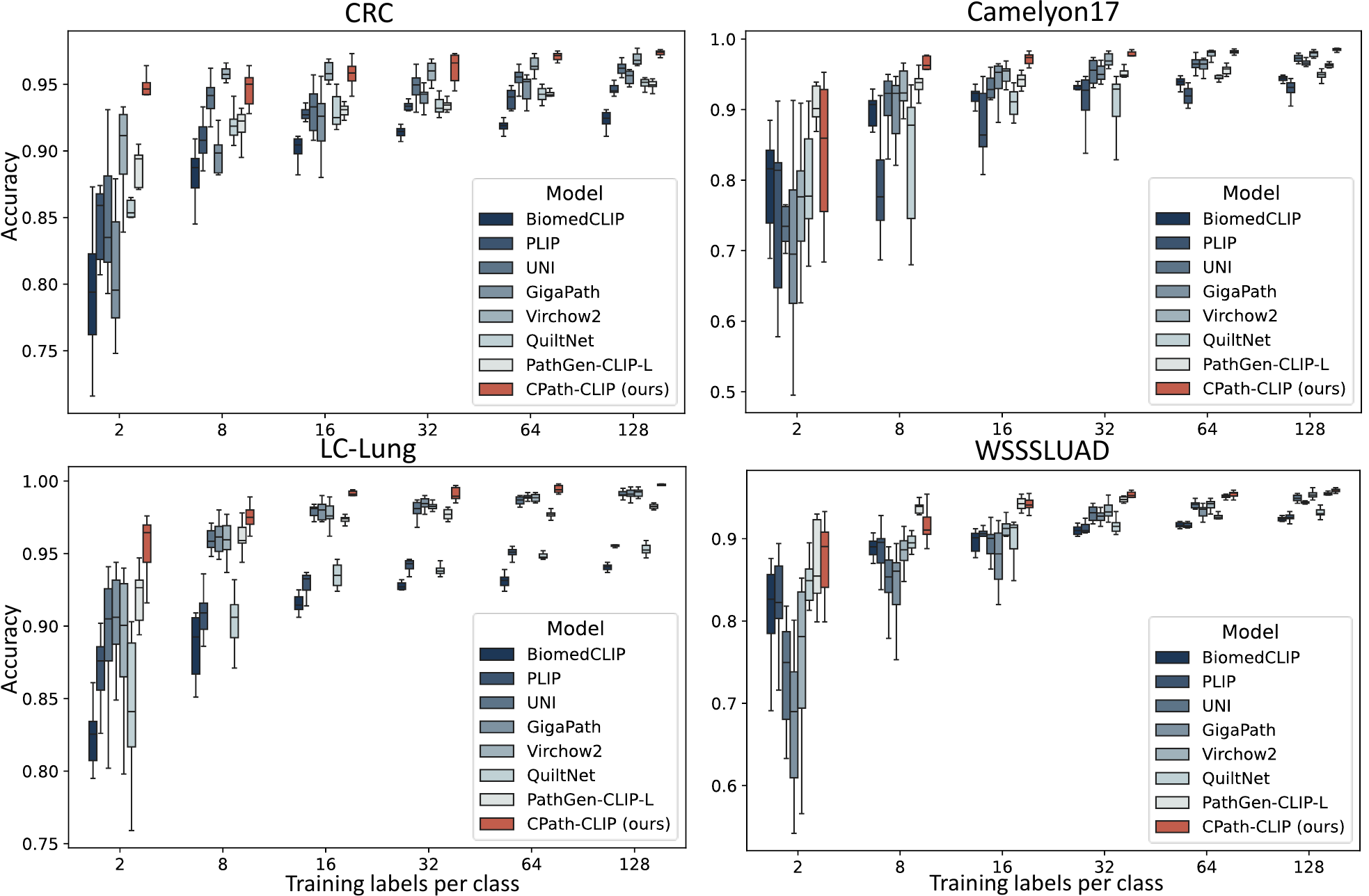}
	
	\caption{Comparison of few-shot classification accuracy (\%) via linear probing across various datasets using different CLIP models.}
	\label{fig:linearprobbox}
\end{figure*}

For few-shot linear probing, we added a fully connected layer to extracted feature representations from four datasets: LC-Colon, Camelyon17, LC-Lung, and WSSSLUAD, using training sizes of 2, 8, 16, 32, 64, and 128 shots. Each size was randomly sampled 10 times and evaluated over 10 runs per configuration. Box plots are utilized to illustrate the variability and robustness of the model's performance.

\noindent\textit{\textbf{Results: CPath-CLIP consistently outperformed previous models.}} As shown in \cref{fig:linearprobbox}, CPath-CLIP demonstrated rapid improvement with minimal data, reaching 95\% accuracy on CRC and LC-Lung using only 2 shots. In contrast, other models achieved less than 91\% accuracy on CRC and approximately 92\% on LC-Lung.

\textit{\textbf{We hypothesize that using advanced LLMs as CLIP's text encoder provides superior world knowledge compared to previous approaches using BERT or GPT-2. When combined with pathology-specific vision model Virchow2, this integration can achieve faster and more effective alignment.}} Unlike general CLIP-based models that require 400 million to billions of training samples, CPath-CLIP was trained with only 700K samples, which holding great potential to redefine future CLIP training paradigms.

\subsection{Benchmarking CPath-Omni at Patch-Level}
At the patch level, we benchmark CPath-Omni against SOTA models, both general-purpose and domain-specific.

We begin by evaluating the performance of various LMMs on VQA using the PathMMU dataset, the largest pathology-specific VQA dataset, which also includes pathologist scores. We compare general-purpose models such as InstructBLIP-FLAN-T5 XXL~\cite{dai2024instructblip}, LLaVA-1.5-13B~\cite{liu2023improvedllava}, Qwen-VL-MAX~\cite{Qwen-VL}, Gemini Pro Vision~\cite{team2023gemini}, and GPT-4V~\cite{openai2023gpt4v}, alongside domain-specific models such as LLaVA-Med, Quilt-LLaVA, and  PathGen-LLaVA.

\noindent\textit{\textbf{Results: CPath-Omni significantly outperforms both the latest pathology-specific model, PathGen-LLaVA, and advanced general-purpose models, even surpassing human-level performance.}} As shown in \cref{tab:overall_results_pathmmu}, CPath-Omni exceeds the current SOTA model, PathGen-LLaVA, by 13.8\%, with a particularly notable 30.1\% improvement in PathCLS~\cite{10729298} performance. Moreover, it slightly surpasses the 71.8\% accuracy achieved by human pathologists, by 0.6\%. We attribute this to the fact that pathologists annotating the PathMMU dataset may not be experts in all disease categories, whereas foundation models like CPath- can learn generalized knowledge across diverse medical fields. This highlights that CPath-Omni has the promising potential of LMMs to offer valuable assistance to clinicians in real-world settings.

\begin{table*}[!t]
	\centering
	\resizebox{\linewidth}{!}{
		\begin{tabular}{@{}lcccccccccccc@{}}
			\toprule
			\textbf{} & \multicolumn{2}{c}{\textbf{Test Overall}} & \multicolumn{2}{c}{\textbf{PubMed}} & \multicolumn{2}{c}{\textbf{SocialPath}} & \multicolumn{2}{c}{\textbf{EduContent}} & \multicolumn{2}{c}{\textbf{Atlas}} &\multicolumn{2}{c}{\textbf{PathCLS}} \\ 
			& Tiny  & ALL  & Tiny  & ALL & Tiny  & All & Tiny  & All  & Tiny  & ALL  & Tiny  & ALL\\
			& (1156)  & (9677)  & (281)  & (3068) & (235)  & (1855) & (255)  & (1938)  & (208)  & (1007) & (177)  & (1809)\\\midrule
			\rowcolor{dino}  Expert performance &  71.8 &  - &  72.9 & -& 71.5 &  - &  69.0 &  - & 68.3 &  - & 78.9 &  - \\
			\midrule
			\multicolumn{13}{c}{\textbf{General Large Multimodal Models}} \\ \midrule
			InstructBLIP-FLAN-T5-XXL & 34.3 & 33.9 & 39.1 & 37.2 & 33.6 & 34.3 & 34.5 & 36.0 & 38.5 & 39.3 & 22.6 & 22.7  \\
			LLaVA-1.5-13B            & 38.8 & 37.6 & 44.5 & 41.0 & 40.4 & 40.4 & 34.1 & 39.4 & 47.1 & 44.3 & 24.9 & 23.5  \\
			Qwen-VL-MAX        & 49.2 & 45.9 & 53.0 & 50.9 & 53.6 & 49.3 & 52.2 & 47.9  & \underline{51.4} & 49.8 & 30.5 & 29.6 \\
			Gemini Pro Vision  & 42.8 & 42.7 & 43.8 & 44.9 & 42.4 & 42.0 & 43.5 & 43.7 & 49.5 & 49.4 & 32.8 & \underline{34.7}   \\
			GPT-4V-1106             & \underline{53.9} & \underline{49.8} & \underline{59.4} & \underline{53.5} & \underline{58.7} & \underline{53.9} & \underline{60.4} & \underline{53.6} & 48.1 & \underline{52.8}  & \underline{36.2} & 33.8\\
			\midrule
			\multicolumn{13}{c}{\textbf{Pathology-specific Large Multimodal Models}} \\ \midrule
			LLaVA-Med & 25.3 & 26.2 & 28.5 & 27.7 & 28.9 & 27.3 & 22.7 & 27.2 & 22.6 & 30.7 & 22.6 & 20.3\\ 			
			Quilt-LLaVA    & 45.6 & 41.5 & 47.3 & 42.6 & 46.4 & 46.6 & 51.8 & 45.3 & 46.2 & 42.7 & 32.2 & 29.2\\
			PathGen-LLaVA & 60.1 & 58.4 & 60.1 & 60.1 & 60.9 & 58.8 & 60.8 & 60.7 & 63.5 & 64.9 & 54.2 & 48.9\\     
			\rowcolor{aliceblue} CPath-Omni & \textbf{72.4} & \textbf{72.2} & \textbf{74.0} & \textbf{69.9} & \textbf{76.6} & \textbf{71.8} & \textbf{69.8} & \textbf{70.6} & \textbf{65.9} & \textbf{70.6} & \textbf{75.7} & \textbf{79.0}\\     
			\bottomrule
	\end{tabular}}
	\caption{Overall results of models on the PathMMU \textbf{test set}. The best-performing LMM in each subset for general and pathology domain LMMs is \textbf{in-bold}, and the top-performing LMM is {\underline{underlined}}.}
	\label{tab:overall_results_pathmmu}
\end{table*}

For the classification task, we evaluate performance across 30 classification datasets, including 16 in-distribution (ID) datasets—where the training data is part of CPath-Omni’s training set—such as VALSET-TCGA, VALSET-CHA, VALSET-WNS \cite{tolkach2023artificial}, Stomach, KIRC, CocaHis~\cite{SITNIK2021102402}, PAIP23, BNCB~\cite{xu2021predicting},  CATCH~\cite{Wilm2022CanineCATCH}, PAIP21, MIDOG22~\cite{MIDDog-dataset}, KICH, CAMEL~\cite{9008367}, Gleason-CNN~\cite{DVN/OCYCMP_2018}, OCELOT~\cite{Ryu_2023_CVPR}, and NASNetLarge~\cite{zoph2018learning}. Additionally, we use 14 out-of-distribution (OOD) datasets that were not included in CPath-Omni’s training data, such as AGGC2022~\cite{AGGC2022-dataset}, KIRP, PAIP19~\cite{PAIP2019}, VALSE-TUKK, as well as 10 datasets from PathCLS within the PathMMU dataset, namely: Skincancer, LC25000-Lung, LC25000-Colon, CRC-100K~\cite{kather2018100}, BACH, WSSSLUAD, PatchCamylon17, Osteo, MHIST~\cite{wei2021petri}, and SICAPv2~\cite{SILVARODRIGUEZ2020105637}. For all 30 datasets, we compare CPath-Omni with the state-of-the-art models GPT-4V and Gemini-1.5-Pro on their respective test sets. For the ID datasets, we also perform task-specific training with Virchow2, the previous SOTA vision-only model, to facilitate direct comparisons with CPath-Omni.

\begin{figure*}[t]
	\centering
	\includegraphics[width=\linewidth]{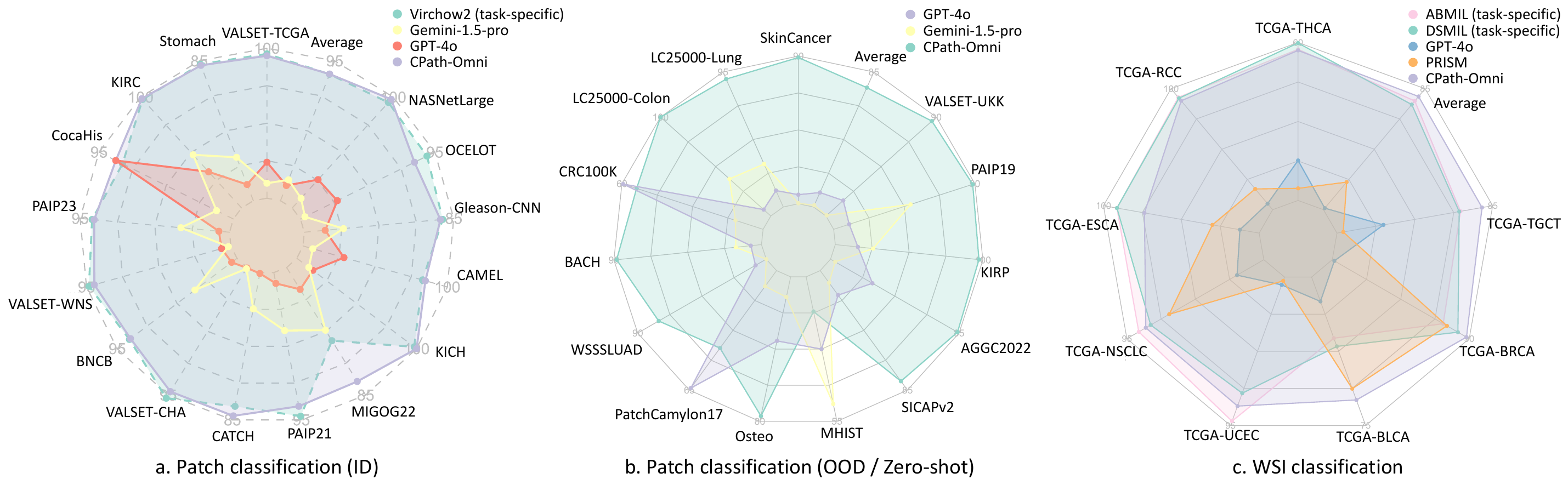}
	
	\caption{Radar plot visualization of CPath-Omni’s performance on patch and WSI classification tasks: (a) patch-level performance under ID conditions, (b) patch classification performance under OOD/zero-shot conditions, and (c) whole-slide image (WSI) performance.}
	\label{fig:cpath_patch_wsi_cls}
\end{figure*}

\noindent\textit{\textbf{Results: CPath-Omni significantly outperforms both GPT-4V and Gemini-1.5-Pro, and even achieves performance comparable to a task-specific fine-tuned version of Virchow2 on each individual dataset.}} As shown in \cref{fig:cpath_patch_wsi_cls}, the radar chart for CPath-Omni closely mirrors that of Virchow2, with performance varying across datasets. However, CPath-Omni holds a slight advantage in average performance. In contrast, for general-purpose models, CPath-Omni consistently outperforms GPT-4V, even on OOD datasets, demonstrating its superior capability and generalization compared to the strongest general-purpose models. Interestingly, when examining the datasets from the PathCLS branch, CPath-Omni’s overall OOD performance is strikingly close to that of CLIP-based models in a zero-shot setting (refer to \cref{tab:zero-shot}). This suggests that CPath-Omni effectively harnesses the power of vision models within its multimodal framework, achieving zero-shot visual classification performance comparable to that of CLIP.

For visual referring prompting and patch captioning, we compare CPath-Omni with domain-specific models such as PathGen-LLaVA, Quilt-LLaVA, and LLaVA-Med. We introduced a set of 50 manually annotated images and conducted both GPT-4V evaluation (by comparing model outputs to ground truth annotations from pathologists) and human evaluation.

\noindent\textit{\textbf{Results: CPath-Omni significantly outperforms the comparison models in both GPT-4V and human evaluations, with the lowest win rate reaching 84\% in the VPR task when compared to Quilt-LLaVA.}} Interestingly, CPath-Omni achieves an even higher win rate against PathGen-LLaVA in the VPR task, despite PathGen-LLaVA being a stronger overall model. We hypothesize that this performance difference arises from the differences in training data: Quilt-LLaVA is trained on videos from YouTube instructors, which may include scenarios resembling visual referring prompting, whereas PathGen-LLaVA is primarily trained on synthetic TCGA data, which lacks such data.

\begin{table}[t]
	\centering
	\resizebox{\columnwidth}{!}{%
		\begin{tabular}{lcccc}
			\toprule
			\multirow{2}{*}{\textbf{CPath-Omni VS.}} & \multicolumn{2}{c}{\textbf{GPT-4o-eval}} & \multicolumn{2}{c}{\textbf{Human-eval}} \\ \cmidrule(lr){2-3} \cmidrule(lr){4-5}
			& \textbf{VPR} & \textbf{Captioning} & \textbf{VPR} & \textbf{Captioning} \\ \midrule
			PathGen-LLaVA  & 96\%  & 82\%  & 98\%  & 80\%  \\
			Quilt-LLaVA    & 84\%  & 96\%  & 90\%  & 92\%  \\
			LLaVA-Med      & 96\%  & 98\%  & 100\% & 100\% \\ 
			\bottomrule
		\end{tabular}%
	}
	\caption{Comparison of CPath-Omni performance on Patch-level VPR and Captioning tasks with GPT-4o-eval and Human-eval across different models. The number represent the percentage of CPath-Omni's responses considered superior.}
	\label{tab:cpath_comparison}
\end{table}
\subsection{Benchmarking CPath-Omni at WSI-Level}
\label{sec:WSI_benchmarking}
In WSI tasks, we benchmark CPath-Omni against both general-purpose models and task-specific fine-tuned models of 3 WSI tasks across 10 datasets.

For the classification task, we evaluate eight TCGA subtyping datasets, including RCC (Kidney Chromophobe, Kidney Renal Clear Cell Carcinoma, Kidney Renal Papillary Cell Carcinoma), NSCLC (Lung Adenocarcinoma, Lung Squamous Cell Carcinoma), BRCA (Invasive Ductal Carcinoma, Invasive Lobular Carcinoma), UCEC (Cystic Mucinous and Serous Neoplasms, Adenomas and Adenocarcinomas), THCA (Papillary Adenocarcinoma, Papillary Carcinoma Columnar Cell, Papillary Carcinoma Follicular Variant), ESCA (Adenomas and Adenocarcinomas, Squamous Cell Neoplasms), BLCA (Transitional Cell Carcinoma, Papillary Transitional Cell Carcinoma), and TGCT (Non-seminoma, Mixed-seminoma, Seminoma). For more detailed information, please refer to the Appendix.

Given that these subtyping tasks are incorporated in CPath-Omni's training, we test on the held-out test set and compare CPath-Omni's performance against task-specific models, including ABMIL~\cite{ITW:2018}, DSMIL~\cite{li2021dual}, a pathology CoCa-style pre-trained model (PRISM), and GPT-4o.

For ABMIL and DSMIL, we use features extracted from WSI patches via CPath-CLIP as input and train the models for 20 epochs, selecting the best checkpoints from the evaluation set for testing on the test set. In PRISM, we use the prompts from the PRISM paper for the eight subtyping tasks it covers. For tasks not included in PRISM, we use the TCGA subtyping classification names as prompts. For GPT-4V, since it does not directly support WSI diagnosis, we first split the WSI into 4096$\times$4096 patches at 20X magnification. GPT-4V then generates descriptions for each patch, and these are merged to form a report for the entire slide. This report is subsequently used as input to prompt GPT-4V for classification of the WSI.

\noindent\textit{\textbf{Results: CPath-Omni significantly outperforms the general-purpose model GPT-4V and the pathology-specific foundation model PRISM, and demonstrates comparable or even superior performance to task-specific fine-tuned models such as ABMIL and DSMIL.}} As shown in \cref{fig:cpath_patch_wsi_cls}, CPath-Omni surpasses ABMIL and DSMIL on three tasks—TCGA-BLCA, TCGA-BRCA, and TCGA-TGCT—and shows a slight overall performance advantage over these models. This suggests that, in the future, a unified framework for WSI classification could achieve the performance of specialized models without the need for task-specific fine-tuning.

For WSI report generation, we compare CPath-Omni with task-specific models, including WsiCaption and HistGen, as well as the general-purpose model PRISM (also supports report generation) and GPT-4o (using the aforementioned method to generate WSI reports). Performance is evaluated using BLEU 1-4 and ROUGE-L scores. For VQA, since the first three models do not support it, we focus the comparison on CPath-Omni and GPT-4o, using the WSI reports generated by GPT-4o as context for answering questions. For closed-ended questions, accuracy is computed, while for open-ended questions, due to the brevity of answers, even small variations or synonyms can cause significant fluctuations in metrics like BLEU and recall. Therefore, we prompt GPT-4o to reference standard answers due to potential variations in BLEU and recall. Note that WSI report generation and VQA tasks do not overlap with data used for WSI classification training.

\noindent\textit{\textbf{Results: CPath-Omni achieves SOTA performance in both WSI report generation and WSI VQA tasks, as shown in \cref{tab:wsi_caption_vqa_comparison}.}} CPath-Omni slightly outperforms the previous SOTA model, HistGen. Note that PRISM tends to generate very short reports (often only a few words expressing the classification), which results in relatively lower performance metrics. In the WSI VQA task, CPath-Omni significantly outperforms GPT-4V, with performance metrics almost doubling: 67.3\% vs. 20.5\% for open-ended questions and 70.8\% vs. 35.5\% for closed-ended questions.

\begin{table}[h]
	\centering
	\large 
	\resizebox{\columnwidth}{!}{%
		\begin{tabular}{@{\hskip 4pt}l@{\hskip 8pt}c@{\hskip 4pt}c@{\hskip 2pt}c@{\hskip 4pt}c@{\hskip 4pt}c|c@{\hskip 5pt}c@{}}
			\toprule
			\textbf{Model}    & \multicolumn{5}{c|}{\textbf{Report Generation}} & \multicolumn{2}{c}{\textbf{VQA}} \\ \midrule
			& \textbf{$BLEU_{1}$} & \textbf{$BLEU_{2}$} & \textbf{$BLEU_{3}$} & \textbf{$BLEU_{4}$} & \textbf{$ROUGE_{L}$} & Open & Closed \\ \midrule
			WSICaption          & 21.8            & 13.7            & 8.6            & 6.4             & 25.1             & -           & -            \\
			HistGen          & 31.8            & 19.7            & 12.7            & 8.4             & 25.4             & -           & -            \\
			PRISM            & 0.0            & 0.0            & 0.0            & 0.0             & 8.0             & -           & -            \\
			GPT-4o           & 15.8            & 6.2             & 2.4            & 0.1             & 12.8             & 20.5           & 35.5            \\
			\rowcolor{aliceblue} CPath-Omni       & \textbf{33.7}            & \textbf{20.1}            & \textbf{12.9}            & \textbf{8.7}             & \textbf{25.6}             & \textbf{67.3}           & \textbf{70.8}            \\ \bottomrule
		\end{tabular}%
	}
	\caption{Comparison of CPath-Omni's performance with task-specific and general models on WSI captioning and VQA tasks.}
	\label{tab:wsi_caption_vqa_comparison}
\end{table}

\section{Conclusion}
In this paper, we present CPath-Omni, a versatile foundational multimodal model designed to tackle both patch-level and WSI-level tasks, spanning captioning, classification, VQA, and visual referring prompting. CPath-Omni's approach enables unified patch and WSI-level training across 30 diverse datasets, allowing knowledge learned from the patch level to simultaneously enhance WSI performance, even trained on a fraction of the data compared to patch-level datasets. Extensive experiments demonstrate that CPath-Omni achieves superior performance across both patch and WSI-level tasks, comparable to or even outperforming task-specific models and significantly surpassing pretrained general-purpose foundation models like PRISM and GPT-4o.  These results highlight the potential of LMMs like CPath-Omni to serve as a ``one-for-all" solution, advancing the next generation of pathology-specific LMMs.
{
    \small
    \bibliographystyle{ieeenat_fullname}
    \bibliography{main}
}
\newpage
\clearpage
\maketitlesupplementary

\setcounter{section}{0}
\renewcommand{\thesection}{\Alph{section}}
\renewcommand{\thesubsection}{\thesection.\arabic{subsection}}
\setcounter{table}{0}
\setcounter{figure}{0}
\renewcommand{\thetable}{\Alph{section}.\arabic{table}}
\renewcommand{\thefigure}{\Alph{section}.\arabic{figure}}

\section{Additional Experiments and Details}

\subsection{Ablations of Vision and Text Components in CPath-CLIP}

We further explore the influence of different vision and text components on CPath-CLIP’s performance in zero-shot tasks to explore its semantic alignment capabilities and understand the role of each element. Our experiments include evaluating CLIP-L alone, Virchow2 alone, and a combination of both, as well as fixing the vision encoder and comparing text encoders by substituting CLIP-L with Qwen2-1.5B. As shown in \cref{tab:zero-shot_ablation}, when using Virchow2 as the fixed vision backbone and replacing the CLIP-L text encoder with Qwen2-1.5B, we observe a 0.9\% overall performance improvement. Conversely, fixing the text encoder as Qwen2-1.5B and replacing CLIP-L with Virchow2 results in a significant 13.7\% performance increase. This suggests that the primary boost is attributed to Virchow2’s pathology-specific pretraining on 3.1 million whole-slide images, highlighting that a more advanced pathology encoder greatly enhances semantic alignment capabilities. Furthermore, combining CLIP-L with Virchow2 provides an additional 0.3\% performance boost. While this gain is modest compared to the standalone Virchow2 encoder, we retain it to enrich semantic features for future integration into LLM.

\subsection{CPath-Omni Performance in Patch and WSI Classification Tasks}
\cref{tab:id_cls}, \cref{tab:ood_cls}, and \cref{tab:wsi_cls} present detailed metrics for patch-level and WSI-level classification corresponding to the radar plot visualization shown in Fig. 4 of the main paper. We also compare state-of-the-art pathology LMMs, Quilt-LLaVA, and PathGen-LLaVA. Notably, these models cannot directly perform WSI classification. To adapt them for this task, we employ the same method used with GPT-4o in the main paper: generating captions for individual patches and merging them into a WSI-level report, which is then used for classification based on predefined questions. Further details are available in Section 5.5 of the main paper.

Our findings show that CPath-Omni significantly outperforms previous models in both patch-level and WSI-level classification tasks. Notably, in out-of-distribution or zero-shot patch classification datasets (\cref{tab:ood_cls}), none of the models were trained on these datasets, making it a relatively fairer comparison. In this context, CPath-Omni significantly surpasses GPT-4o and Gemini-1.5-Pro, the strongest general models, as well as pathology-specific LMMs such as PathGen-LLaVA and Quilt-LLaVA. Additionally, CPath-Omni achieves performance comparable to task-specific fine-tuned models, underscoring its strength in task unification and exceptional overall performance.

\subsection{Experiment Details of Patch-level Linear Probing and Task-Specific Model Fine-Tuning}
\textbf{Linear Probing:} The linear probing experiment evaluates the representational power of a pre-trained model by adding a linear layer to its output. This linear layer maps the model's output vector to the number of classes, enabling classification. The experiment uses a batch size of 32 and runs for 20 epochs. The optimizer is AdamW with a learning rate of \(1 \times 10^{-2}\). To ensure robustness and reproducibility, we employ 10 different seeds. The procedure involves randomly selecting N samples (2, 8, 16, 32, 64, 128) from each class to form the training set. If an official test set is unavailable or lacks labels, the remainder of the dataset serves as the test set. Throughout the 20 epochs, we select the best-performing model based on its accuracy on the test set, providing insights into the effectiveness of the added linear layer in classifying unseen data.

\noindent\textbf{Task-specific model fine-tuning:} For task-specific model fine-tuning, we build on the linear probing setup by unfreezing the Virchow2 backbone and performing full-parameter fine-tuning on the model using the entire training set.

\subsection{Details for WSI preprocessing}
For WSI preprocessing, we utilize CLAM to identify and segment regions by setting appropriate thresholds. Within each WSI, we extract 2048 $\times$ 2048 non-overlapping patches at a magnification of 40$\times$ from the identified regions. Patches are retained if more than 10\% of their area contains valid tissue regions. Additionally, each 2048 $\times$ 2048 patch is further subdivided into one 2048 $\times$ 2048 patch, four 1024 $\times$ 1024 patches, and sixteen 512 $\times$ 512 patches for subsequent feature extraction.

\subsection{Details for WSI Task-Specific Fine-Tuning}
The models are trained for 20 epochs without a learning rate schedule, using a fixed learning rate of $1 \times 10^{-5}$. The training process utilizes the Adam optimizer without weight decay, and the batch size is consistently set to 1. 

\noindent{\textbf{Model Architecture.}}
The MIL framework commonly used for WSI classification includes three learnable components:
(1) A fully-connected layer to reduce the dimensionality of features to 256.
(2) An attention network to aggregate and transform the instance features.
(3) A final fully-connected layer for making predictions. We experiment with ABMIL and DSMIL. Both models share the same fully connected layers for reducing feature dimensionality and making predictions. For the attention network, ABMIL uses the gated attention mechanism, while DSMIL introduces a dual-stream architecture. 

For a fair comparison, the input patch features of these two models are kept consistent with those of CPath-Omni.

\subsection{Training Hyperparameters for CPath-CLIP and CPath-Omni}
We detail the training parameters of CPath-CLIP in  \cref{tab:hparams_cpath_clip}. The hyperparameters for the four training stages of CPath-Omni are listed in \cref{tab:hparams_stage_1}, \cref{tab:hparams_stage_2}, \cref{tab:hparams_stage_3}, and \cref{tab:hparams_stage_4}, respectively. Specifically, stages 1 and 2 focus on patch-based training, stage 3 is dedicated to WSI-based training, and stage 4 involves a mix of patch-based and WSI-based training.

\subsection{Hardware}
\label{appendix:subsec:hardware}

We employ 8 NVIDIA H800-80G GPUs to train the CPath-Omni model, 1 NVIDIA A100-40G GPU for fine-tuning task-specific models, and 4 NVIDIA H800-80G GPUs for caption generation using PathGen-LLaVA and Quilt-LLaVA.

\section{Additional Details of Collected Datasets}

\begin{figure}[t]
	\centering
	\includegraphics[width=0.7\linewidth]{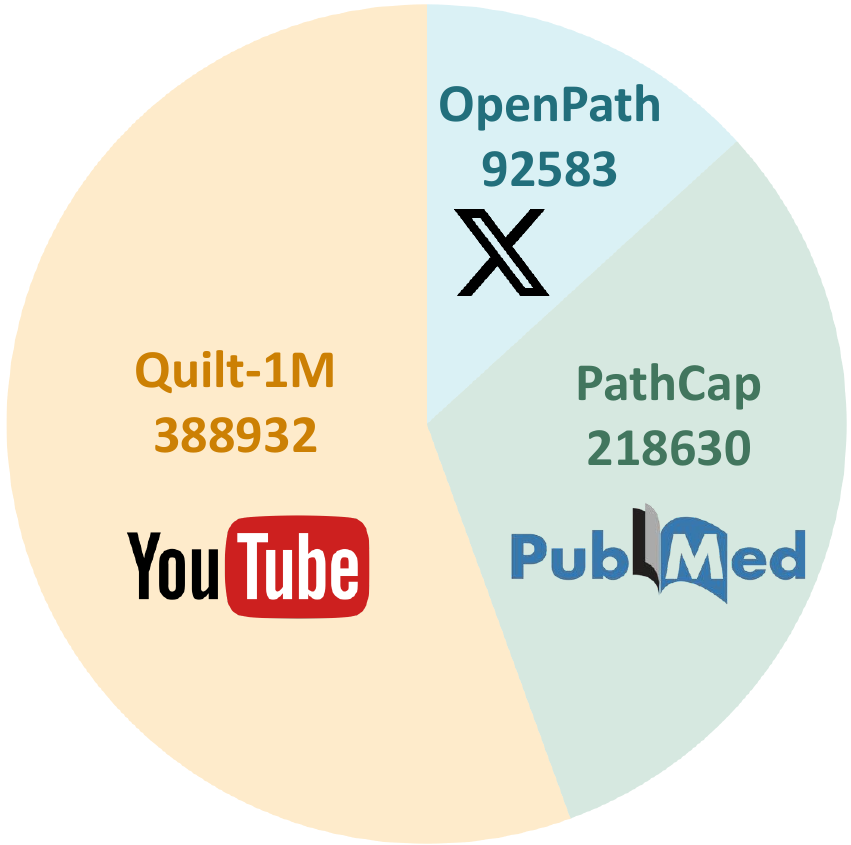}
	
	\caption{CProportions of sub-datasets in CPath-PatchCaption and their primary sources.}
	\label{fig:cpath_patchcaption}
\end{figure}

\begin{figure*}[t]
	\centering
	\includegraphics[width=0.8\linewidth]{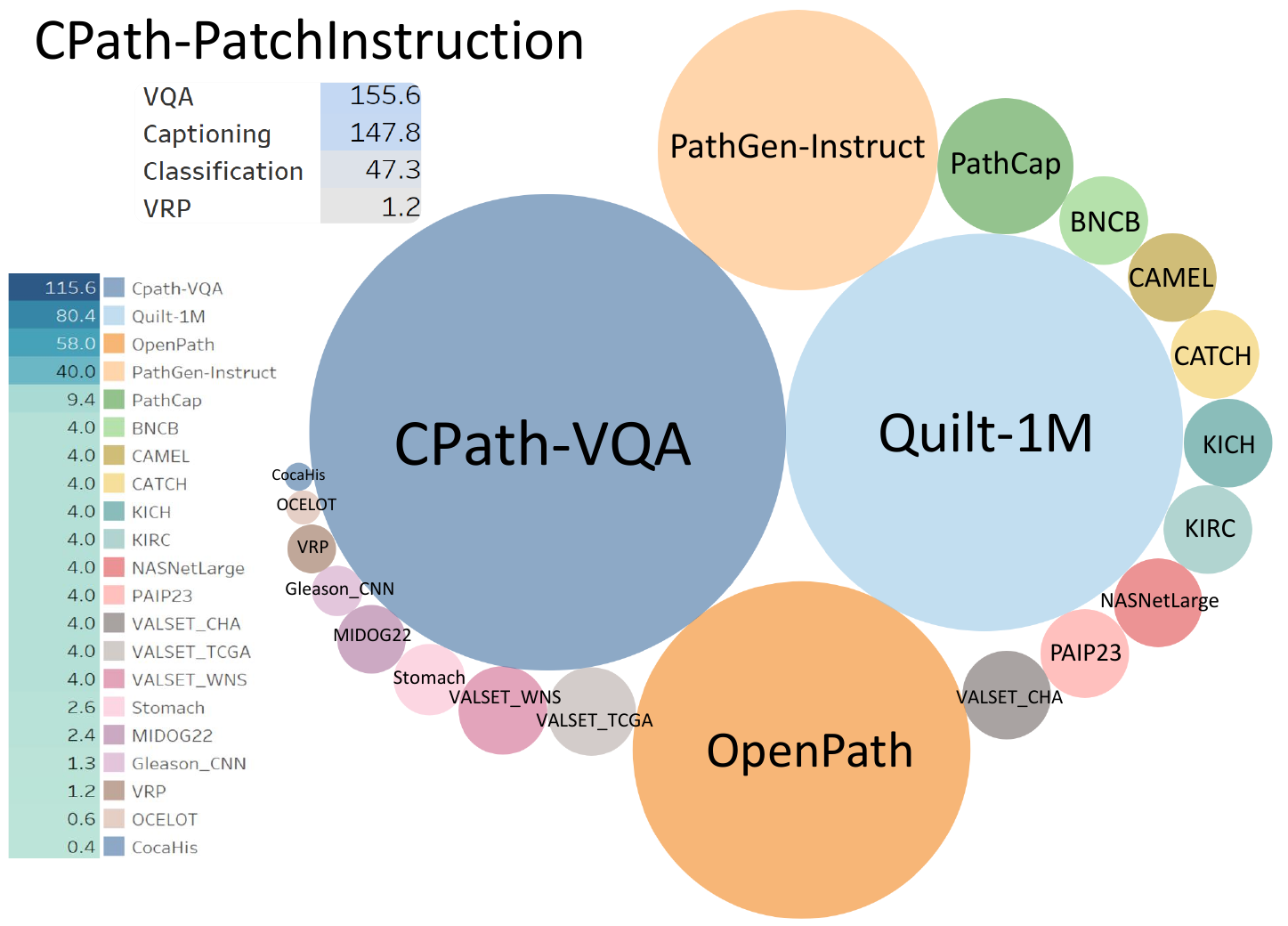}
	
	\caption{Visualization of the datasets used in CPath-PatchInstruct, including their quantities (in thousands) and proportional distributions, where larger circles represent higher proportions.}
	\label{fig:cpath_patchinstruct}
\end{figure*}

\begin{figure*}[t]
	\centering
	\includegraphics[width=0.8\linewidth]{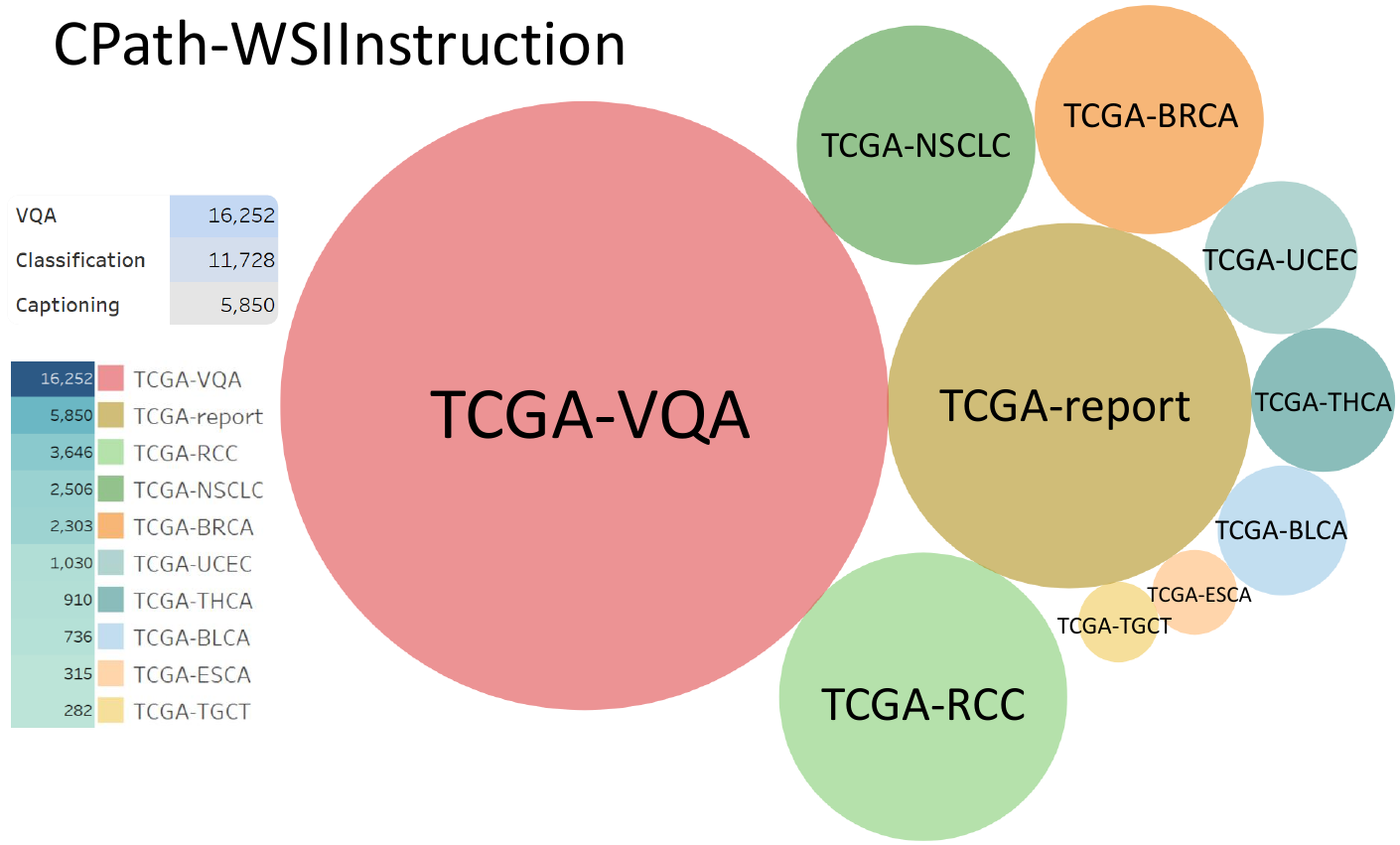}
	
	\caption{Visualization of the datasets used in CPath-WSIInstruct, including their quantities and proportional distributions, where larger circles represent higher proportions.}
	\label{fig:cpath_wsiinstruct}
\end{figure*}

We provide details on the dataset sources and distribution of CPath-PatchCaption in \cref{fig:cpath_patchcaption}. The sources, quantities, distributions, and sub-task data allocations for CPath-PatchInstruction and CPath-WSIInstruction are also illustrated in \cref{fig:cpath_patchinstruct} and \cref{fig:cpath_wsiinstruct}.

Additionally, the construction of CPath-VQA within the CPath-Instruct dataset follows a systematic approach. First, we collect datasets that already include captions. For datasets lacking captions, such as classification datasets, we use GPT-4o to generate captions by combining classification labels with image data. GPT-4o further generates VQA pairs from these captioning datasets, creating the CPath-VQA.

\begin{figure*}[t]
	\centering
	\includegraphics[width=0.95\linewidth]{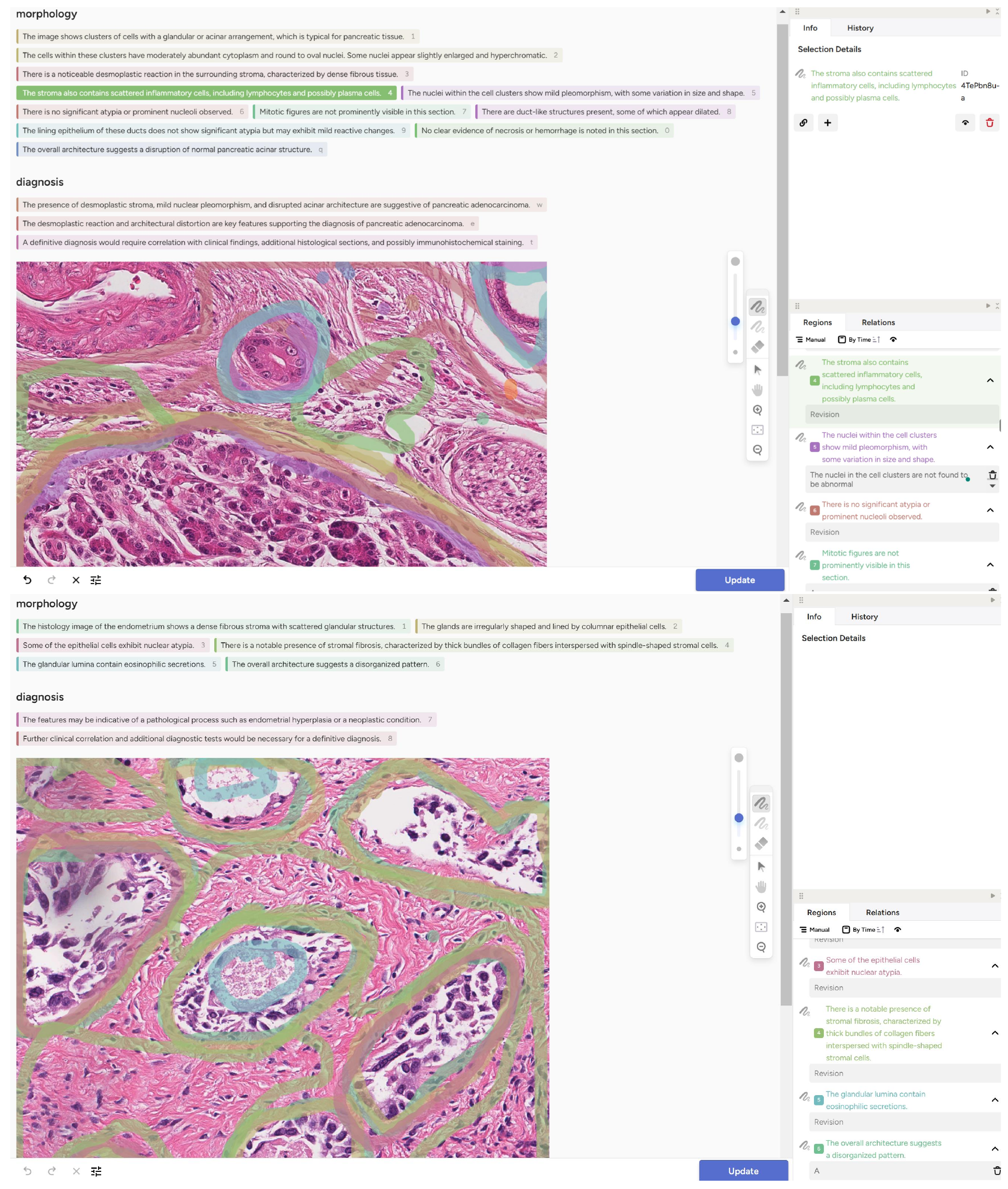}
	
	\caption{Examples of the pathologist annotation interface for visual referring prompting. Pathologists are required to verify whether the given morphology and diagnosis are correct, record "T" or "F" in the bottom-right corner, or modify the original findings as needed. Once all findings are confirmed accurate, they use corresponding colored markers to highlight the regions in the image associated with each finding.}
	\label{fig:vrp_annotation_examples}
\end{figure*}

\begin{figure*}[t]
	\centering
	\includegraphics[width=0.9\linewidth]{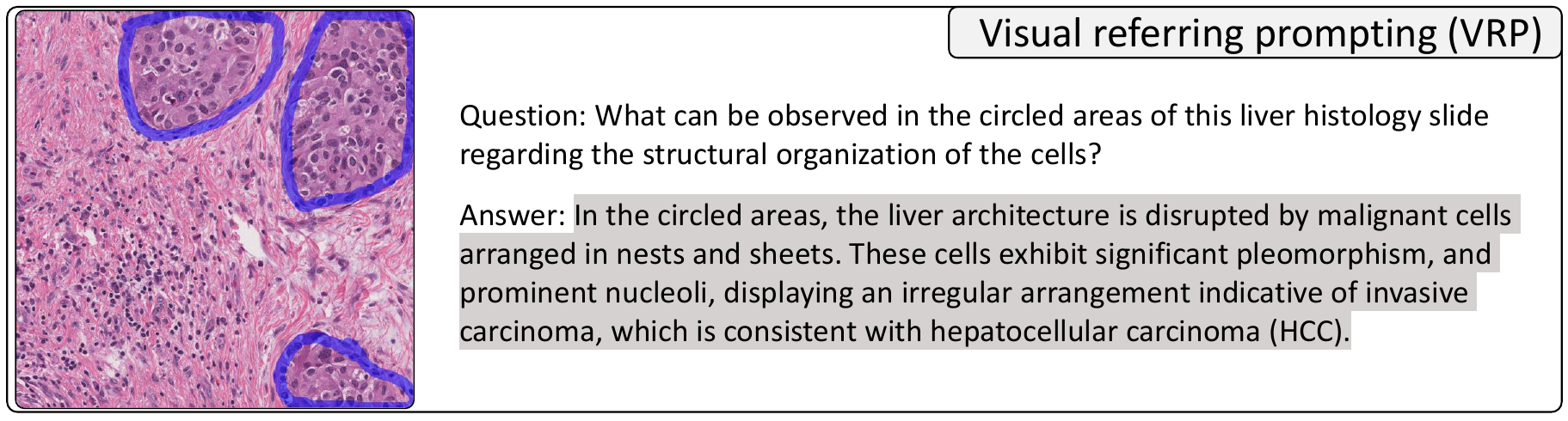}
	
	\caption{An example of a constructed visual referring prompting task, where questions are answered based on the highlighted regions.}
	\label{fig:VRP}
\end{figure*}

\begin{figure*}[t]
	\centering
	\includegraphics[width=0.9\linewidth]{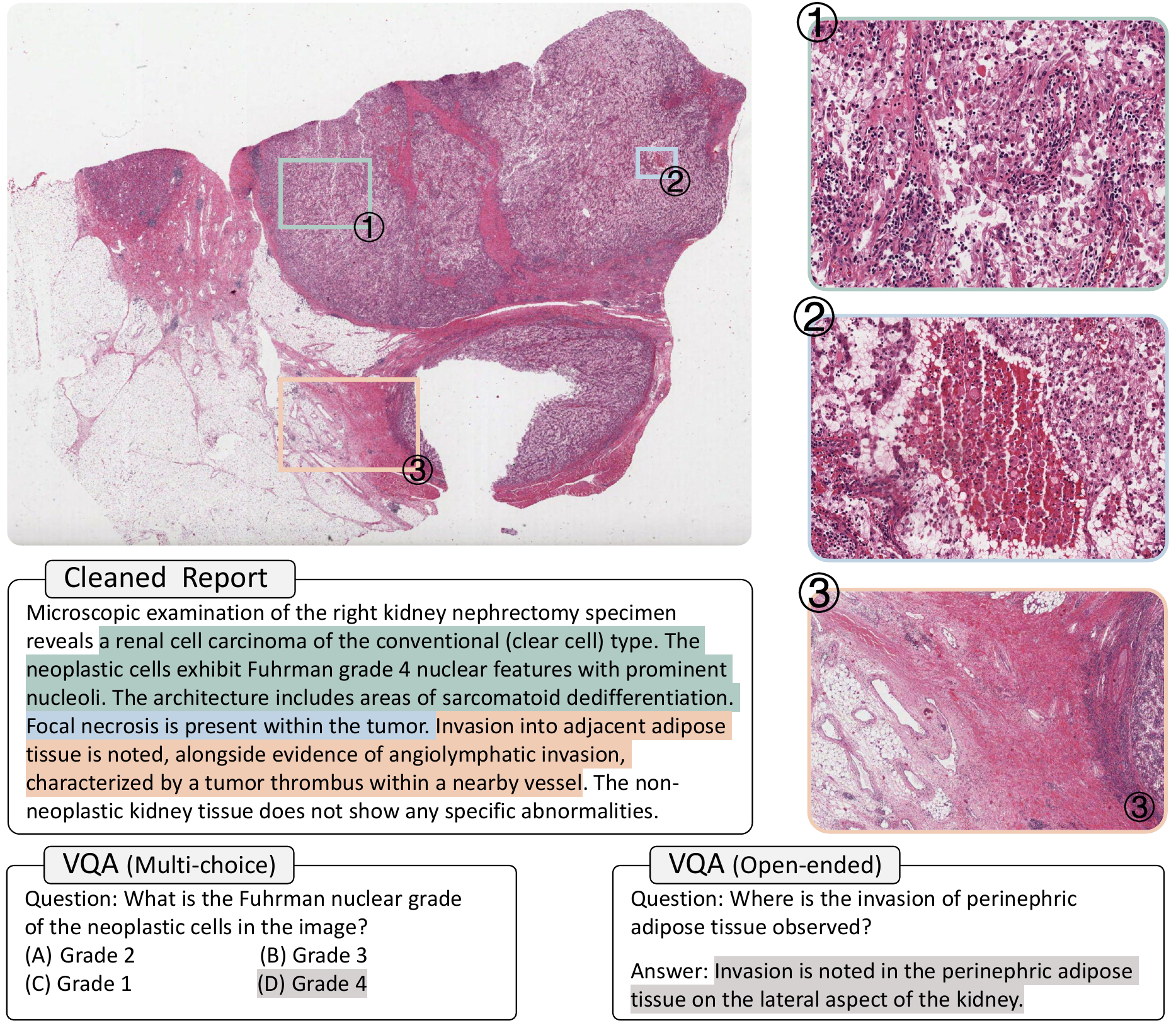}
	
\caption{Examples of WSI Captioning and VQA tasks, where corresponding findings in the captions are highlighted with matching colored boxes in the WSI.}
\label{fig:wsi_question_type_examples}
\end{figure*}

We also present visualization examples of the novel task of visual referring prompting, alongside tasks specific to whole-slide images, which differ from natural images due to their extremely high resolution, reaching nearly 100,000 $\times$ 100,000 pixels. Specifically, \cref{fig:vrp_annotation_examples} illustrates the annotation interface for visual referring prompting and shows how pathologists annotate this task. \cref{fig:VRP} provides an example of generating visual referring prompting. \cref{fig:wsi_question_type_examples} showcases examples of WSI captioning and VQA. In these examples, we present the cleaned report, which, after processing, aligns well with the pathological representations in the whole-slide image. We highlight the correspondences between features in the report and the whole-slide image for better clarity. Based on this cleaned report, we generate both multiple-choice and open-ended examples.

\begin{table*}[ht]
	
	\centering
	\resizebox{0.95\linewidth}{!}{
		\begingroup
		\renewcommand{\arraystretch}{1.2}
		\large
		\begin{tabular}{@{}ccccccccccccc@{}}
			\toprule
			\multicolumn{2}{c}{\textbf{Vision Encoder}} & \multirow{2}{*}{\textbf{Text Encoder}} & \multirow{2}{*}{\textbf{LC-Lung}} & \multirow{2}{*}{\textbf{LC-Colon}} & \multirow{2}{*}{\textbf{CRC100K}} & \multirow{2}{*}{\textbf{SkinCancer}} & \multirow{2}{*}{\textbf{Pcam}} & \multirow{2}{*}{\textbf{BACH}} & \multirow{2}{*}{\textbf{Osteo}} & \multirow{2}{*}{\textbf{WSSSLUAD}} & \multirow{2}{*}{\textbf{SICAPv2}} & \multirow{2}{*}{\textbf{Average}} \\
			\textbf{CLIP-L} & \textbf{Virchow2} &  &  &  &  &  &  &  &  &  &  &  \\
			\midrule
			\XSolidBrush & \Checkmark & CLIP-L & 94.9 & 99.5 & 77.7 & 70.1 & 91.1 &\textbf{74.5} & 72.7 & \textbf{89.9} & \textbf{67.9} & 82.0 \\
			\XSolidBrush & \Checkmark & Qwen2-1.5B & 96.8 & 99.8 & 76.0 & \textbf{78.1} & 94.6 & 72.0 & \textbf{82.0} & 86.4 & 60.7 & 82.9 \\
			\Checkmark & \XSolidBrush & Qwen2-1.5B & 92.1 & 91.9 & 67.0 & 61.7 & 87.8 & 47.5 & 52.9 & 79.0 & 42.9 & 69.2 \\
			\rowcolor{aliceblue} \Checkmark & \Checkmark & Qwen2-1.5B & \textbf{97.1} & \textbf{100.0} & \textbf{78.0} & 74.2 & \textbf{95.9} & 72.3 & 80.7 & 87.1 & 63.1 & \textbf{83.2} \\
			\bottomrule
		\end{tabular}
		\endgroup}
	\caption{Zero-shot classification performance comparison of CPath-CLIP built with different vision and text encoders.}
	\label{tab:zero-shot_ablation}
\end{table*}

\begin{table*}[t]

	\centering
		\resizebox{0.8\linewidth}{!}{
	\begin{tabular}{lccccc>{\columncolor{aliceblue}}c}
		\toprule
		& GPT-4o & Gemini-1.5-pro & Quilt-LLaVA & PathGen-LLaVA & Full-finetune & CPath-Omni (ours) \\
		\midrule
		VALSET\_TCGA & 39.2 & 28.0 & 23.5 & 28.9 & \textbf{97.0} & \underline{96.0} \\
		Stomach & 18.7 & 33.3 & 21.6 & 19.6 & \textbf{83.2} & \underline{82.6} \\
		KIRC & 79.8 & 84.4 & 38.2 & 90.8 & \underline{99.4} & \textbf{99.6} \\
		CocaHis & \textbf{90.0} & 60.0 & 43.4 & 89.7 & 88.0 & \textbf{90.0} \\
		PAIP23 & 30.0 & 47.8 & 14.5 & 15.9 & \textbf{89.4} & \underline{88.4} \\
		VALSET\_WNS & 29.5 & 26.2 & 17.3 & 23.3 & \textbf{93.8} & \underline{91.2} \\
		BCNB & 52.0 & 65.8 & 65.7 & 65.2 & \textbf{90.4} & \underline{90.0} \\
		VALSET\_CHA & 30.2 & 30.6 & 16.2 & 25.2 & \textbf{96.6} & \underline{93.2} \\
		CATCH & 20.4 & 36.0 & 26.7 & 26.7 & \underline{79.0} & \textbf{83.2} \\
		PAIP21 & 7.0 & 37.4 & 18.9 & 18.1 & \textbf{92.6} & \underline{86.2} \\
		MIDOG22 & 47.7 & 62.1 & 50.9 & 48.9 & \underline{65.8} & \textbf{80.2} \\
		KICH & 74.0 & 73.0 & 29.6 & 87.1 & \underline{99.4} & \textbf{100.0} \\
		CAMEL & 64.2 & 53.4 & 56.7 & 61.0 & \underline{91.4} & \textbf{92.4} \\
		Gleason\_CNN & 46.3 & 51.8 & 38.8 & 39.0 & \textbf{81.7} & \textbf{81.7} \\
		OCELOT & 44.2 & 27.3 & 20.9 & 37.2 & \textbf{90.9} & \underline{84.4} \\
		NASNetLarge & 62.8 & 54.4 & 50.4 & 54.5 & \underline{97.6} & \textbf{99.0} \\
		\midrule
		Average & 46.0 & 48.2 & 33.3 & 45.7 & \underline{89.8} & \textbf{89.9} \\
		\bottomrule
	\end{tabular}}
	\caption{Performance comparison of general-purpose, pathology-specific, and task-specific models on ID patch classification tasks.}
	\label{tab:id_cls}
\end{table*}

\begin{table*}[ht]
	\centering

			\resizebox{0.7\linewidth}{!}{
	\begin{tabular}{lcccc>{\columncolor{aliceblue}}c}
		\toprule
		& GPT-4o & Gemini-1.5-pro & Quilt-LLaVA & PathGen-LLaVA & CPath-Omni (ours)  \\
		\midrule
		SkinCancer & 33.7 & 30.2 & 6.7 & 42.4 & \textbf{89.4} \\
		LC25000-Lung & 46.7 & 57.5 & 63.2 & \underline{79.8} & \textbf{92.1} \\
		LC25000-Colon & 81.3 & 87.5 & 92.1 & \textbf{100.0} & \textbf{100.0} \\
		CRC100K & \textbf{59.8} & 39.9 & 16.2 & 57.1 & \underline{57.4} \\
		BACH & 29.7 & 36.3 & 15.3 & \underline{42.1} & \textbf{88.8} \\
		WSSSLUAD & \underline{62.5} & 60.0 & 42.1 & 43.9 & \textbf{85.1} \\
		PatchCamylon17 & \underline{64.4} & 34.6 & 58.1 & \textbf{66.2} & 52.7 \\
		Osteo & \underline{63.3} & 54.2 & 33.6 & 34.9 & \textbf{78.9} \\
		MHIST & \underline{50.0} & \textbf{53.8} &  \underline{50.0} & \underline{50.0} & 47.4 \\
		SICAPv2 & \underline{41.3} & 35.6 & 25.7 & 26.9 & \textbf{80.9} \\
		AGGC2022 & \underline{51.3} & 36.8 & 18.1 & 27.4 & \textbf{84.4} \\
		KIRP & 74.6 & 77.7 & 59.7 & \underline{92.2} & \textbf{99.2} \\
		PAIP19 & 54.4 & \underline{71.8} & 33.4 & 43.1 & \textbf{89.4} \\
		VALSET\_UKK & \underline{39.4} & 30.4 & 23.1 & 19.1 & \textbf{87.6} \\
			\midrule
		Average & \underline{53.7} & 50.5 & 38.4 & 51.8 & \textbf{81.0} \\
		\bottomrule
	\end{tabular}}
\caption{Performance comparison of general-purpose and pathology-specific models on OOD patch classification tasks.}
		\label{tab:ood_cls}
\end{table*}

\begin{table*}[t]
	\centering
			\resizebox{0.8\linewidth}{!}{
	\begin{tabular}{lcccccc>{\columncolor{aliceblue}}c}
		\toprule
		& ABMIL & DSMIL & GPT-4o & PRISM & Quilt-LLaVA & PathGen-LLaVA & CPath-Omni (ours) \\
		\midrule
		TCGA-THCA & \underline{58.7} & \textbf{59.9} & 37.6 & 32.3 & 29.5 & 40.5 & 58.5 \\
		TCGA-RCC  & \textbf{95.7} & \underline{95.3} & 43.0 & 50.3 & 38.6 & 51.1 & 94.0 \\
		TCGA-ESCA & \textbf{97.4} & \textbf{97.4} & 73.7 & 79.0 & 63.2 & 76.3 & 92.1 \\
		TCGA-NSCLC & \textbf{91.1} & 87.2 & 58.8 & 81.1 & 58.8 & 65.5 & \underline{88.8} \\
		TCGA-UCEC & \textbf{93.4} & 83.0 & 42.9 & 41.4 & 47.0 & 47.4 & \underline{87.8} \\
		TCGA-BLCA & 60.3 & 61.7 & 54.1 & \underline{68.8} & 51.4 & 53.7 & \textbf{70.7} \\
		TCGA-BRCA & 82.4 & \underline{86.7} & 50.6 & 83.5 & 48.1 & 60.5 & \textbf{89.2} \\
		TCGA-TGCT & \underline{72.6} & 72.1 & 42.9 & 27.4 & 20.5 & 39.3 & \textbf{80.9} \\
		\midrule
		Average & \underline{81.5} & 80.4 & 50.5 & 58.0 & 44.6 & 54.3 & \textbf{82.8} \\
		\bottomrule
	\end{tabular}}
\caption{Performance comparison of general-purpose and pathology-specific models on WSI classification tasks using balanced accuracy.}
			\label{tab:wsi_cls}
\end{table*}

\begin{table*}[h!]
	\centering
	\renewcommand{\arraystretch}{1.5}
	\setlength{\tabcolsep}{10pt}
	\begin{tabular}{|p{3cm}|p{11cm}|}
		\hline
		\textbf{Dataset} & \textbf{Classes} \\
		\hline
		PatchCamelyon & 
		\textit{`lymph node', `lymph node metastasis'} \\
		\hline
		NCK-CRC & 
		\textit{`Adipose', `Debris', `Lymphocytes', `Mucus', `Smooth muscle', `Normal colon mucosa', `Cancer-associated stroma', `Colorectal adenocarcinoma epithelium'} \\
		\hline
		LC25000Lung & 
		\textit{`Lung adenocarcinoma', `benign lung tissue', `lung squamous cell carcinomas'} \\
		\hline
		LC25000Colon & 
		\textit{`Colon adenocarcinoma', `normal colon tissue'} \\
		\hline
		BACH & 
		\textit{`Benign tissue', `In-situ carcinoma', `Invasive carcinoma', `Normal tissue'} \\
		\hline
		SICAPv2 & 
		\textit{`Non-cancerous', `Atrophic well differentiated and dense glandular regions', `Cribriform, ill-formed, large-fused and papillary glandular patterns', `Isolated cells or file of cells, nests of cells without lumina formation and pseudo-rosetting patterns'} \\
		\hline
		Osteo & 
		\textit{`Non-tumor', `Necrotic tumor', `Viable tumor'} \\
		\hline
		SkinCancer & 
		\textit{`Non-tumor chondral tissue', `Non-tumor dermis', `Non-tumor elastosis', `Non-tumor epidermis', `Non-tumor hair follicle', `Non-tumor skeletal muscle', `Non-tumor necrosis', `Non-tumor nerves', `Non-tumor sebaceous glands', `Non-tumor subcutis', `Non-tumor sweat glands', `Non-tumor vessel', `Tumor epithelial basal cell carcinoma', `Tumor epithelial squamous cell carcinoma', `Tumor melanoma', `Tumor naevus'} \\
		\hline
		WSSSLUAD & 
		\textit{`tumor', `normal'} \\
		\hline
	\end{tabular}
	\caption{Classes for each dataset in zero-shot image classification. Consistent prompt templates are used for all datasets, including: 
		\texttt{`An H\&E image of \{\}'}, \texttt{`This is an image of \{\} presented in the image'}, and \texttt{`An H\&E patch of \{\}'}.}
	\label{tab:prompts}
\end{table*}

\begin{table*}[]
	\centering
	\begin{tabular}{p{5.5cm}|p{3.5cm}}
		\toprule
		Hyper-parameter & Value \\
		\midrule
		Num GPUs & 8 \\
		Num epochs & 5 \\
		Learning rate & 3e-5 \\
		Per device train batch size & 64 \\
		Gradient accumulation steps & 1 \\
		Weight decay & 0.1 \\
		Warmup steps & 300 \\
		\bottomrule
	\end{tabular}
	\caption{Hyperparameters used in CPath-CLIP training.}
	\label{tab:hparams_cpath_clip}
\end{table*}

\begin{table}[th]
	\centering
	\begin{tabular}{p{3.8cm}|p{3.2cm}}
		\toprule
		Hyper-parameter & Value \\
		\midrule
		LLM Model & Qwen2.5-14B-Instruct \\
		Vision Model & CPath-CLIP \\
		Tunable parts & MLP \\
		Vision select layer & -2 \\
		Model max length & 8192 \\
		\midrule
		Image aspect ratio & Square \\
		Image grid pinpoints & None \\
		Patch merge type & Flat \\
		Prompt version & Plain \\
		\midrule
		Num GPUs & 8 \\
		Num epochs & 1 \\
		Learning rate & 1e-3 \\
		Per device train batch size & 16 \\
		Gradient accum steps & 1 \\
		Weight decay & 0. \\
		Warmup ratio & 0.03 \\
		Lr scheduler type & cosine \\
		\bottomrule
	\end{tabular}
	\caption{Hyperparameters used in stage 2 training of CPath-Omni (Patch pretraining).}
	\label{tab:hparams_stage_1}
\end{table}

	\begin{table}[th]
		\centering
		\begin{tabular}{p{4cm}|p{3.5cm}}
			\toprule
			Hyper-parameter & Value \\
			\midrule
			LLM Model & Qwen2.5-14B-Instruct \\
			Vision Model & CPath-CLIP \\
			Tunable parts & CPath-CLIP \& MLP \& LLM \\
			Vision select layer & -2 \\
			Model max length & 32768 \\
			\midrule
			Image aspect ratio & AnyRes (up to 9 splits) \\
			Image grid pinpoints & (1x1),...,(3x3) \\
			Patch merge type & Spatial unpad \\
			Prompt version & Qwen-1.5 \\
			\midrule
			Num GPUs & 8 \\
			Num epochs & 1 \\
			Learning rate & 1e-5 \\
			Per device train batch size & 1 \\
			Gradient accum steps & 8 \\
			Weight decay & 0. \\
			Warmup ratio & 0.03 \\
			Lr scheduler type & cosine \\
			Vision tower lr & 2e-6 \\
			\bottomrule
		\end{tabular}
		\caption{Hyperparameters used in stage 2 training of CPath-Omni (Patch fine-tuning).}
		\label{tab:hparams_stage_2}
	\end{table}

\begin{table}[th]
	\centering
	\begin{tabular}{p{3.8cm}|p{3.2cm}}
		\toprule
		Hyper-parameter & Value \\
		\midrule
		LLM Model & Qwen2.5-14B-Instruct \\
		Vision Model & CPath-CLIP \\
		Tunable parts & WSI projector \\
		Vision select layer & -2 \\
		Model max length & 8192 \\
		WSI hidden size & 3328 \\
		\midrule
		Image aspect ratio & Square \\
		Image grid pinpoints & None \\\
		Patch merge type & Flat \\
		Prompt version & Plain \\
		\midrule                 		
		Num GPUs & 8 \\
		Num epochs & 1 \\
		Learning rate & 5e-6 \\
		Per device train batch size & 16 \\
		Gradient accum steps & 1 \\
		Weight decay & 0. \\
		Warmup ratio & 0.1 \\
		Lr scheduler type & cosine \\
		\bottomrule
	\end{tabular}
	\caption{Hyperparameters used in stage 3 training of CPath-Omni (WSI pretraining).}
	\label{tab:hparams_stage_3}
\end{table}
\begin{table}[th]
	\centering
	\begin{tabular}{p{3.9cm}|p{3.4cm}}
		\toprule
		Hyper-parameter & Value \\
		\midrule
		LLM Model & Qwen2.5-14B-Instruct \\
		Vision Model & CPath-CLIP \\
		Tunable parts & CPath-CLIP \& MLP \& WSI projector \& LLM \\
		Vision select layer & -2 \\
		Model max length & 32768 \\
		Wsi hidden size & 3328 \\
		\midrule
		Image aspect ratio & AnyRes (up to 9 splits) \\
		Image grid pinpoints & (1x1),...,(3x3) \\
		Patch merge type & Spatial unpad \\
		Prompt version & Qwen-1.5 \\
		\midrule
		Num GPUs & 8 \\
		Num epochs & 5 \\
		learning rate & 1e-5 \\
		Per device train batch size & 1 \\
		Gradient accum steps & 8 \\
		Weight decay & 0. \\
		Warmup ratio & 0.1 \\
		Lr scheduler type & cosine \\
		WSI projector lr & 1e-5 \\
		Vision tower lr & 2e-6 \\
		\bottomrule
	\end{tabular}
	\caption{Hyperparameters used in stage 4 training of CPath-Omni (mixed patch and WSI fine-tuning).}
	\label{tab:hparams_stage_4}
\end{table}

\begin{table*}[]
	\centering

	\resizebox{\textwidth}{!}{
		\begin{tabular}{p{3cm} p{14cm}}
			\hline
			\textbf{Dataset} & \textbf{Source Link} \\
			\hline
			TCGA & \raggedright\url{https://portal.gdc.cancer.gov/} \tabularnewline
			CocaHis & \raggedright\url{https://portal.gdc.cancer.gov/} \tabularnewline
			BCNB & \raggedright\url{https://bcnb.grand-challenge.org/} \tabularnewline
			CAMELYON17 & \raggedright\url{https://camelyon17.grand-challenge.org/Data/} \tabularnewline
			MIDOG2022 & \raggedright\url{https://midog.deepmicroscopy.org/download-dataset/} \tabularnewline
			AGGC2022 & \raggedright\url{https://aggc22.grand-challenge.org/} \tabularnewline
			ARCH & \raggedright\url{https://warwick.ac.uk/fac/cross_fac/tia/data/arch} \tabularnewline
			BACH & \raggedright\url{https://zenodo.org/records/3632035} \tabularnewline
			CAMEL & \raggedright\url{https://drive.google.com/open?id=1brr8CnU6ddzAYT157wkdXjbSzoiIDF9y} \tabularnewline
			LC2500 & \raggedright\url{https://academictorrents.com/details/7a638ed187a6180fd6e464b3666a6ea0499af4af} \tabularnewline
			MIDOG2021 & \raggedright\url{https://imig.science/midog2021/download-dataset/} \tabularnewline
			OCELOT & \raggedright\url{https://zenodo.org/record/7844149} \tabularnewline
			Osteo & \raggedright\url{https://www.cancerimagingarchive.net/collection/osteosarcoma-tumor-assessment/} \tabularnewline
			PAIP2019 & \raggedright\url{https://paip2019.grand-challenge.org/} \tabularnewline
			PAIP2020 & \raggedright\url{https://paip2020.grand-challenge.org/} \tabularnewline
			PAIP2021 & \raggedright\url{https://paip2021.grand-challenge.org/} \tabularnewline
			SICAPv2 & \raggedright\url{https://data.mendeley.com/datasets/9xxm58dvs3/1} \tabularnewline
			CRC-100K & \raggedright\url{https://zenodo.org/records/1214456} \tabularnewline
			PCam & \raggedright\url{https://github.com/basveeling/pcam} \tabularnewline
			HistGen & \raggedright\url{https://github.com/dddavid4real/HistGen} \tabularnewline
			PathGen-Instruct & \raggedright\url{https://github.com/PathGen-1-6M/PathGen-1.6M} \tabularnewline
			PathCap &
			\raggedright\url{https://huggingface.co/datasets/jamessyx/PathCap} \tabularnewline
			OpenPath &
			\raggedright\url{https://drive.google.com/drive/folders/1b5UT8BzUphkHZavRG-fmiyY9JWYIWZER} \tabularnewline
			Quilt-1M &
			\raggedright\url{https://github.com/wisdomikezogwo/quilt1m} \tabularnewline
			PathMMU &
			\raggedright\url{https://huggingface.co/datasets/jamessyx/PathMMU} \tabularnewline
			CocaHis &
			\raggedright\url{https://www.sciencedirect.com/science/article/abs/pii/S1746809420305085?via\%3Dihub} \tabularnewline
			OCELOT &
			\raggedright\url{https://ocelot2023.grand-challenge.org/datasets} \tabularnewline
			Gleason\_CNN &
			\raggedright\url{https://github.com/eiriniar/gleason_CNN} \tabularnewline
			MIDOG22 &
			\raggedright\url{https://midog2022.grand-challenge.org} \tabularnewline
			VALSET  &
			\raggedright\url{https://zenodo.org/records/7548828} \tabularnewline
			PAIP23  &
			\raggedright\url{https://2023paip.grand-challenge.org/} \tabularnewline
			NASNetLarge  &
			\raggedright\url{https://zenodo.org/records/3825933} \tabularnewline
			RCdpia (KIRC, KICH, KIRP)  &
			\raggedright\url{http://39.171.241.18:8888/RCdpia/annotation.php} \tabularnewline
			CATCH  &
			\raggedright\url{https://www.cancerimagingarchive.net/collection/catch/} \tabularnewline
			SkinCancer  &
			\raggedright\url{https://heidata.uni-heidelberg.de/dataset.xhtml?persistentId=doi:10.11588/data/7QCR8S} \tabularnewline
			MHIST  &
			\raggedright\url{https://bmirds.github.io/MHIST} \tabularnewline
			WSSSLUAD  &
			\raggedright\url{https://wsss4luad.grand-challenge.org/}
			 \tabularnewline
			LC25000  &
			\raggedright\url{https://github.com/tampapath/lung_colon_image_set?tab=readme-ov-file}
			\tabularnewline

			\hline
		\end{tabular}
	}
	\caption{\textbf{Datasets used in this study with corresponding access links}}
	\label{tab:data_links}
	
\end{table*}

\section{Prompts for GPT-4o}
\begin{figure*}[t]
	\centering
	\includegraphics[width=0.9\linewidth]{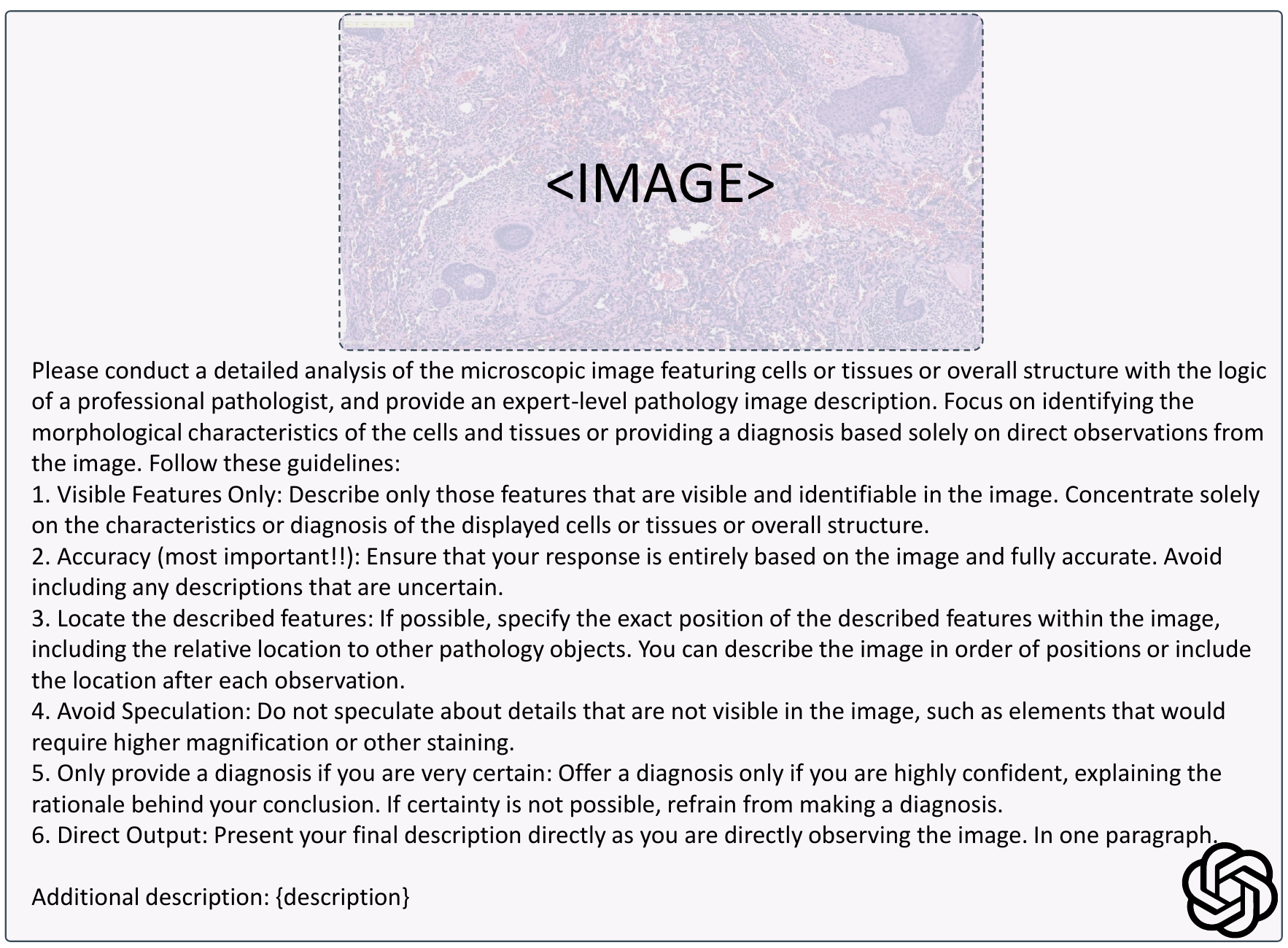}
	\caption{Prompt for GPT-4o to generate a detailed description for an image based on its original caption.}
	\label{fig:prompt_description_generation}
\end{figure*}

\begin{figure*}[t]
	\centering
	\includegraphics[width=0.9\linewidth]{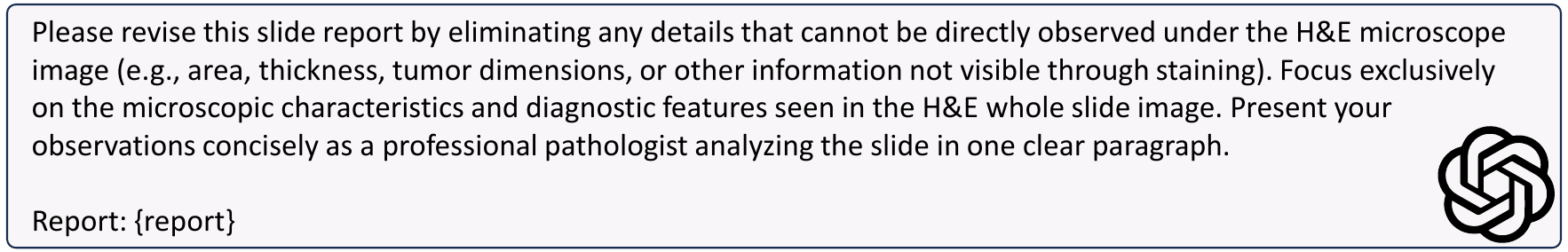}
	\caption{Prompt for GPT-4o to clean the raw data from the WSI report, transforming it into accurate ground truth WSI report.}
	\label{fig:wsi_report_revise}
\end{figure*}

\begin{figure*}[t]
	\centering
	\includegraphics[width=0.9\linewidth]{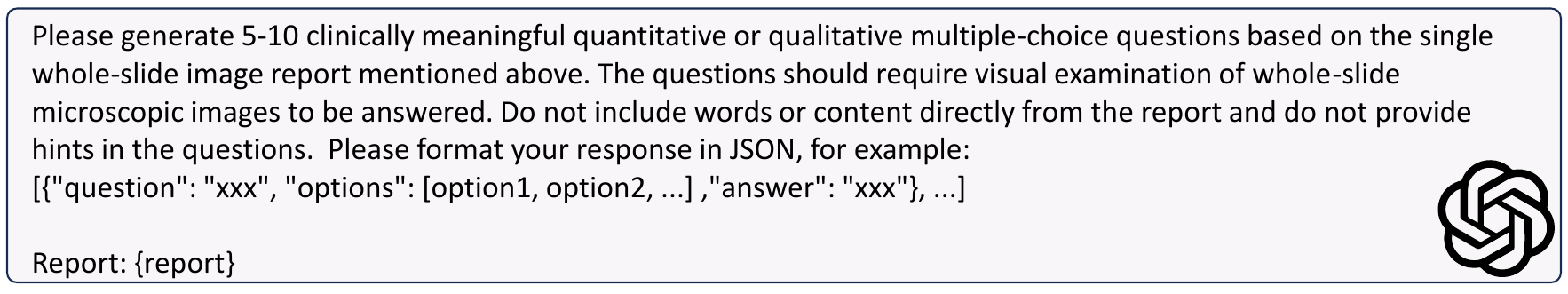}
	\caption{Prompt for GPT-4o to generate closed-ended VQA based on a given WSI report.}
	\label{fig:report_vqa_multichoice}
\end{figure*}

\begin{figure*}[t]
	\centering
	\includegraphics[width=0.9\linewidth]{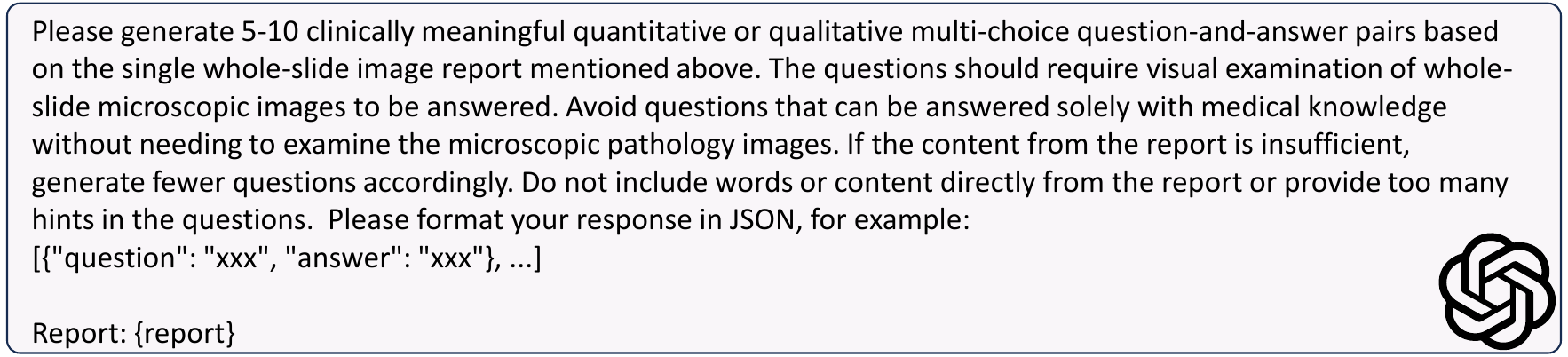}
	\caption{Prompt for GPT-4o to generate open-ended VQA based on a given WSI report.}
\label{fig:report_vqa_openended}
\end{figure*}

\begin{figure*}[t]
	\centering
	\includegraphics[width=0.9\linewidth]{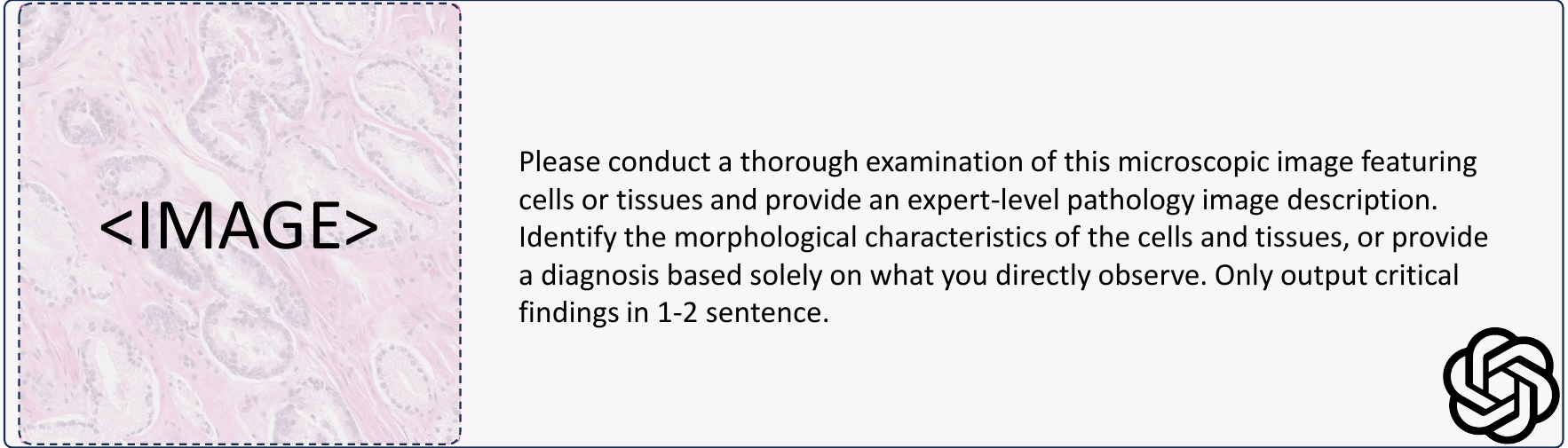}
	\caption{Prompt for GPT-4o to generate the caption for WSI patch image.}
	\label{fig:gpt4o_wsi_caption_generation}
\end{figure*}

\begin{figure*}[t]
	\centering
	\includegraphics[width=0.9\linewidth]{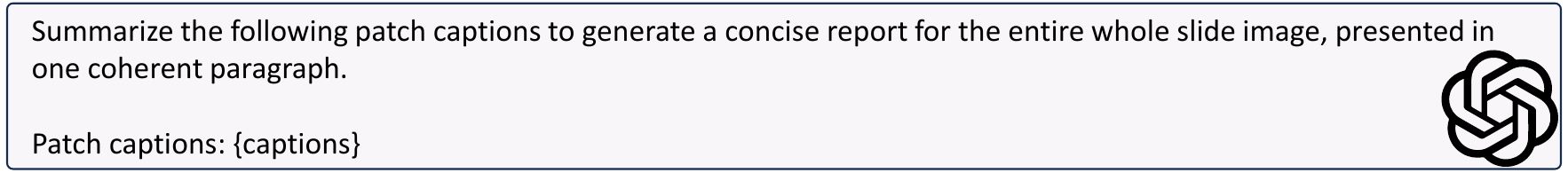}
	\caption{Prompt for GPT-4o to merge generated patch captions into a comprehensive WSI report.}
	\label{fig:merge_caption_to_report}
\end{figure*}

\begin{figure*}[t]
	\centering
	\includegraphics[width=0.9\linewidth]{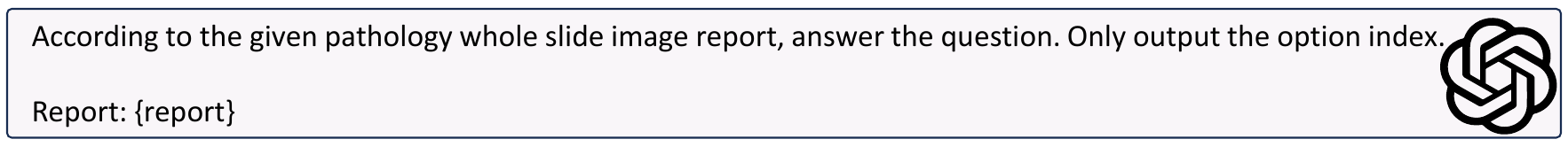}
	\caption{Prompt for GPT-4o to answer the closed-ended question based on the generated WSI report.}
	\label{fig:report_to_answer}
\end{figure*}

\begin{figure*}[t]
	\centering
	\includegraphics[width=0.9\linewidth]{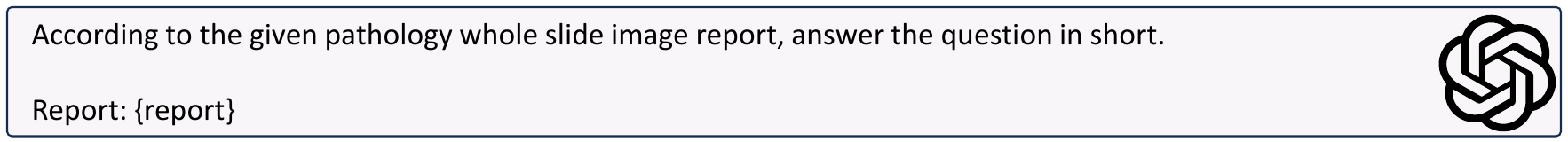}
	\caption{Prompt for GPT-4o to answer the open-ended question based on the generated WSI report.}
	\label{fig:report_to_answer_open}
\end{figure*}

\begin{figure*}[t]
	\centering
	\includegraphics[width=0.9\linewidth]{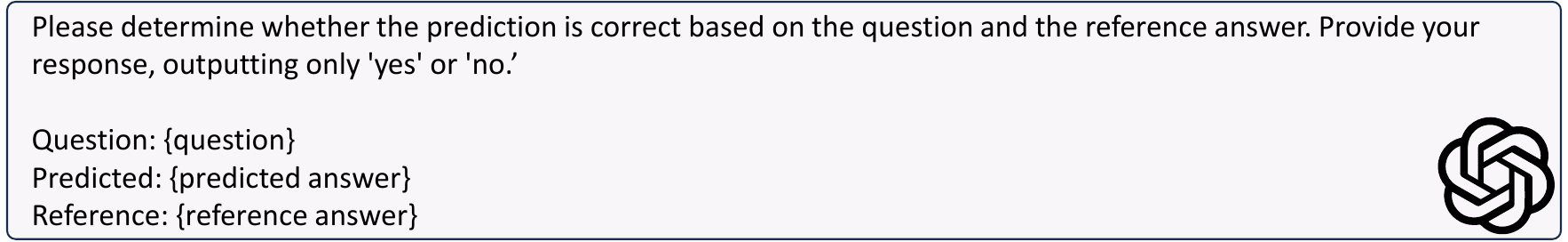}
	\caption{Prompt for GPT-4o to determine whether the predicted answer to the open-ended question is correct.}
	\label{fig:judger}
\end{figure*}

This section presents all the prompts used in our dataset and experimental process, including: (1) the prompt in Figure \cref{fig:prompt_description_generation}, which is used with GPT-4o to enrich and refine existing image captions by adding details; (2) the prompt in Figure \cref{fig:wsi_report_revise}, which is utilized to modify raw WSI reports by removing information that cannot be directly observed in the WSI, such as gross specimen descriptions; (3) the prompts in \cref{fig:report_vqa_multichoice} and \cref{fig:report_vqa_openended}, which are applied to generate closed-ended and open-ended VQA pairs based on cleaned WSI reports; (4) the prompt in \cref{fig:gpt4o_wsi_caption_generation}, which is designed to prompt GPT-4o to generate captions for image patches within a WSI; (5) the prompt in \cref{fig:merge_caption_to_report}, which is used to merge all the captions generated for individual patches in a WSI into a cohesive generated WSI report; (6) the prompts in \cref{fig:report_to_answer} and \cref{fig:report_to_answer_open}, which are employed to guide GPT-4o in answering closed-ended and open-ended WSI VQAs by analyzing whether the answers can be derived from the generated WSI report; and (7) the prompt in Figure \ref{fig:judger}, which is used to determine whether the answer to an open-ended WSI VQA is correct by referencing the provided question and answer.

\end{document}